\pdfoutput=1

\documentclass[11pt]{article}

\usepackage{acl}
\usepackage{times}
\usepackage{latexsym}

\usepackage[T1]{fontenc}

\usepackage[utf8]{inputenc}

\usepackage{microtype}

\usepackage{inconsolata}

\usepackage{graphicx}
\usepackage{graphicx,subcaption,booktabs,float}
\usepackage{booktabs, tabularx}
\usepackage{booktabs}
\usepackage[dvipsnames]{xcolor}
\definecolor{owl}{RGB}{219, 48, 122}

\usepackage{amsmath}

\title{
Causal Understanding by LLMs: The Role of Uncertainty}

\author{
 \textbf{Oscar Lithgow-Serrano\textsuperscript{1}\thanks{Equal contribution}},
 \textbf{Vani Kanjirangat\textsuperscript{1}\footnotemark[1]},
 \textbf{Alessandro Antonucci\textsuperscript{1}}
\\
 \textsuperscript{1}SUPSI, IDSIA, Switzerland\\ 
\\
}

\begin{document}
\maketitle

    

\begin{abstract}

Recent papers show LLMs achieve near-random accuracy in causal relation classification, raising questions about whether such failures arise from limited pretraining exposure or deeper representational gaps.
We investigate this under uncertainty-based evaluation, testing whether pretraining exposure to causal examples improves causal understanding using >18K PubMed sentences—half from \emph{The Pile} corpus, half post-2024—across seven models (Pythia-1.4B/7B/12B, GPT-J-6B, Dolly-7B/12B, Qwen-7B). We analyze model behavior through: (i) \emph{causal classification}, where the model identifies causal relationships in text, and (ii) \emph{verbatim memorization probing}, where we assess whether the model prefers previously seen causal statements over their paraphrases. Models perform four-way classification (direct/conditional/correlational/no-relationship) and select between originals and their generated paraphrases. 
Results show 
almost identical accuracy on seen/unseen sentences (p>0.05), no memorization bias (24.8\% original selection),  output distribution over the possible options almost flat --- with entropic values near the maximum (1.35/1.39), confirming random guessing. Instruction-tuned models show severe miscalibration (Qwen: >95\% confidence, 32.8\% accuracy, ECE=0.49). Conditional relations induce highest entropy (+11\% vs direct). These findings suggest that failures in causal understanding arise from the lack of structured causal representation, 
rather than insufficient exposure to causal examples during pretraining.

\end{abstract}

\section{Introduction}


Causal understanding from text, intended here as the ability of an LLM to identify whether a text includes a statement about a causal relation between two entities, and which is the causal direction of such a relation, is a critical task for modern natural language understanding. Previous work demonstrates that \emph{Large Language Models} (LLMs) struggle with such causal tasks, achieving near-random performance on benchmarks requiring causal inference 
\cite{Ashwani2024, Feng2024b, Guo2017, Joshi2024a, Kanjirangat2024}. Recent works showed the importance of analyzing the underlying model uncertainty to achieve better results, or at least to understand the reasons for poor performances
\cite{Cui2025, Shorinwa2025}. From this perspective, a very promising direction is provided by distinguishing between different sources of uncertainty, such as \emph{epistemic}, corresponding to the uncertainty related to lack of knowledge about the underlying model, and \emph{aleatoric}, that is the intrinsic ambiguity of the process
\cite{Hullermeier2021}. Another crucial aspect is how the presence of \emph{seen versus unseen} data — i.e., content observed during pretraining or \textit{familiar} observations — affects uncertainty and model behavior in terms of causal understanding. While uncertainty quantification in LLMs has been explored in prior work 
\cite{He, Liu2025d, Yadkori2024}, 
the link between uncertainty sources and familiar causal patterns in the context of causal understanding remains underexamined.

We design controlled experiments to examine how these uncertainty sources arise in the context of causal understanding. 
We consider \emph{memorization} as one of the tasks to understand the effect of seen \emph{verbatim} causal patterns, in line with the uncertainty sources. Using scientific conclusion sentences from PubMed abstracts, we test whether models trained on these exact texts (via \emph{The Pile} dataset, \citet{Gao2020}) show reduced uncertainty compared to similar but unseen texts and, if this has an impact on their accuracy.

Our approach uses two complementary tests for memorization effects. First, if models truly understood causal patterns from training (not just surface forms), they should exhibit substantially lower uncertainty and higher accuracy on familiar data (i.e., training data) versus unseen data.
Second, if only surface memorization occurred, we would expect substantial differences in uncertainty between original sentences and their paraphrases within the training dataset.
In a nutshell, we use uncertainty metrics to explore memorization effects in causal understanding and whether this reflects representational limitations rather than data exposure.

\textbf{Our contributions:} We propose uncertainty-based quantification (i.e., entropy, ECE/ACE) as a way for analyzing LLMs' causal understanding  competence, linking calibration metrics to performance on causal tasks.
We conduct experiments to examine how uncertainty sources, including memorization of verbatim causal statements, influence causal understanding in LLMs.
Within this framework, we show:
(i) Exposure to training data does not guarantee memorization or improved performance — models show identical accuracy on seen versus unseen texts;
(ii) Identify overconfidence in most performing models, with high confidence predictions despite very low accuracy; 
(iii) Quantify that conditional causal relationships induce the highest uncertainty, suggesting models lack nuanced causal representations; 
(iv) In the experiments, we consider two datasets, an existing one from the literature and an extension constructed by us to be used for testing memorization and causal understanding.

Our experimental findings indicate that uncertainty in LLM causal understanding reflects epistemic limitations rather than insufficient exposure to training examples. Models do not leverage already observed patterns for causal tasks, instead exhibiting systematic uncertainty that correlates with task complexity rather than data familiarity.

\section{Related Work}

Recent work examines uncertainty sources in LLMs. \citet{Kirchhof2025a} demonstrates that models can assess their uncertainty through verbalized confidence. \citet{Giulianelli2023} proposes semantic entropy to measure uncertainty in free-form generation. However, these approaches focus on general tasks rather than structured reasoning. \citet{Wang2024r} shows LLMs struggle with calibrated uncertainty in knowledge-intensive tasks, consistent with our findings in causal reasoning.

Memorization's role in LLM capabilities remains debated. \citet{Carlini2021} demonstrate that models memorize training data verbatim, while \cite{Li2024f, Zhang2023i} show this memorization can be beneficial. \citet{Tirumala2022} quantifies memorization across model scales. Other findings show that memorization alone cannot explain model capabilities, requiring 100+ exact repetitions for reliable recall \cite{Kandpal2023, Li2024f}.

While prior work evaluates causal understanding in LLMs through various benchmarks, none have examined it through the lens of uncertainty sources and verbatim memorization recalls.  Our work extends this by showing memorization fails to improve structured reasoning tasks. Our approach uniquely combines controlled exposure to training data with uncertainty quantification, revealing that causal understanding requires more than pattern memorization, especially when the complexity increases.

The rest of the paper is organized as follows. In Section \ref{sec:data}, we present the datasets used for the experimental study and analyses. Section \ref{sec:method} presents the detailed discussion of the proposed uncertainty-based quantification, and the experimental setup is presented in Section \ref{sec:exp}. Results and analysis are reported in Section \ref{sec:result}, with detailed discussion in Section \ref{sec:discuss}.

\section{Data Construction}\label{sec:data}
First, we used two datasets of sentences labeled with their causal types to test the impact of exposure to causal patterns during pretraining on the accuracy and uncertainty of causal understanding tasks with LLMs.

We used \citet{Yu2019}'s dataset, consisting of 3,061 sentences from science findings classified into four causal relationship types: direct causal, conditional causal, correlational, or no relationship. 
The original dataset lacked source abstracts for the extracted sentences. 
As these were needed to train 
a classifier for extending the dataset with more recent abstracts, 
we searched PubMed using near-exact sentence matches. This yielded a filtered dataset with the following distribution: direct causal (234), conditional causal (113), correlational (489), and no relationship (598). This resulting dataset is hereafter referred to as \textit{Original}.
Although filtering removed almost half of the entries, the label distribution remains highly similar to the original (total variation distance = $0.026$).
Future work will consider more sophisticated matching schemes to preserve more data.

To control for verbatim memorization, we created an extension from PubMed abstracts published after 2024, beyond our models' training cutoff. We use a BERT-based classifier (F1-score = 0.97) trained on the original annotations to label 5,400 new sentences, then subsample to match the class distribution of our \textit{Original} dataset. This resulting dataset is hereafter referred to as \textit{Newer}


We complement the datasets by generating, with GPT-4o-mini, for each sentence: (i) five paraphrases especially focused on preserving the causal relationship; 
(ii) one negation that reverses the causal relationship; (iii) two questions that probe understanding of the causal content. Considering the original plus paraphrase resulted in 18,366 sentences. 

Since both datasets are used in \emph{Multiple Choice Question Answering} (MCQA) setups, we hereafter refer to the \textit{original}-based dataset as MCQA and the \textit{newer} one as MCQA-newer. For instance, the samples generated for the Causal Type Classification Task are depicted in Figure \ref{fig:task1}. Examples from the generated dataset can be found in Appendix \ref{app:data}. 
\floatstyle{boxed}
\restylefloat{figure}
\begin{figure*}[htp!]
\centering
\tiny
\begin{verbatim}
{"qa_idx": 0, "context": "However, the small sample size in this study limits its generalizability to diverse populations, 
so we call for future research that explores SSL-powered personalization at a larger scale.",

"text": "However, the small sample size in this study limits its generalizability to diverse populations, 
so we call for future research that explores SSL-powered personalization at a larger scale.", 
"text_type": 0, "causal_class_label": 0, 

"choices": [{"label": 1, "text": "Direct Causal", "description": "The statement explicitly states that one variable directly causes changes in another."}, 
{"label": 3, "text": "Correlational", "description": "The statement describes an association between variables, but no causation is 
explicitly stated."}, 
{"label": 2, "text": "Conditional Causal", "description": "The statement suggests causation but includes uncertainty through hedging words or
modal expressions."}, 
{"label": 0, "text": "No Relationship", "description": "No correlation or causation relationship is mentioned."}, 
{"label": 4, "text": "Other", "description": ""}]

\end{verbatim}
\caption{Examples from the constructed data (Task 1) - Casual Type Classification}\label{fig:task1}
\end{figure*}



\section{Uncertainty Quantification}\label{sec:method}
We focus on LLMs answering multiple-choice questions. Let $\mathcal{Y}$ denote the set of possible options and $Y$ the corresponding variable. The probability distribution over the possible choices $P(Y)$ is assumed to be available. 
We quantify uncertainty through multiple metrics.

\textbf{Entropy.} We can describe the model uncertainty related to this task by the entropy \cite{shannon1948mathematical}, i.e., $H:=-\sum_{y \in \mathcal{Y}} P(y) \ln P(y)$. This is a non-negative function taking the value of zero for deterministic distributions, and its maximum value for uniform distributions. In the case of quaternary variables, the value of the maximum is 1.39.

\textbf{Calibration.} We bin predictions by confidence level and compute actual accuracy within each bin. Perfect calibration yields a diagonal relationship between confidence and accuracy. We then compute \emph{Expected Calibration Error} (ECE) and \emph{Adaptive Calibration Error} (ACE) \cite{nixon2019measuring, posocco2021estimating}.



ECE measures how well a model's estimated probabilities match the observed probabilities. A perfectly calibrated model has zero ECE. It is computed as the weighted average of the absolute differences between average accuracy and average confidence.
\begin{equation}
\mathrm{ECE} = \sum_{r=1}^{R} \frac{|B_r|}{n} \left| \mathrm{acc}(B_r) - \mathrm{conf}(B_r) \right|
\end{equation}

\noindent
Where,
 \( R \) is the number of bins (typically fixed-width over the interval \([0, 1]\)),
  \( B_r \) is the set of indices of predictions with confidence scores in the \( r \)-th bin,
 \( n \) is the total number of samples,
  \( \mathrm{acc}(B_r) = \frac{1}{|B_r|} \sum_{i \in B_r} \mathbf{1}(\hat{y}_i = y_i) \) is the accuracy in bin \( r \),
\( \mathrm{conf}(B_r) = \frac{1}{|B_r|} \sum_{i \in B_r} \hat{p}_i \) is the average confidence in bin \( r \),
\( \hat{p}_i = \max_{k} p_i^{(k)} \) is the predicted confidence for sample \( i \) and  \( k \) is the number of labels/classes.

 To overcome the limitations of ECE, such as the bias-variance trade-off induced by binning approaches and its alignment to binary-class settings \cite{Guo2017}, ACE was proposed, which utilizes flexible binning \cite{Nixon2019}. ACE is motivated by the bias-variance trade-off, which suggests that an effective estimate of overall calibration error should emphasize regions where predictions are concentrated, while placing less weight on sparsely populated regions. ACE takes as input the predictions \emph{P}, correct labels, and a number of ranges \emph{R}:
\begin{equation}
\mathrm{ACE} = \frac{1}{R} \sum_{r=1}^{R} \left| \mathrm{acc}(B_r) - \mathrm{conf}(B_r) \right|
\end{equation}

\noindent
Where,
  \( R \) is the number of bins (adaptively chosen so each bin contains roughly the same number of samples),
  \( B_r \), \( \mathrm{acc}(B_r) \), \( \mathrm{conf}(B_r) \), and \( \hat{p}_i \) are defined as above.

\textbf{Consistency.} For sentences with multiple paraphrases, we measure whether models make consistent predictions across semantically equivalent inputs.

\textbf{Statistical Tests.} We apply chi-square tests for original versus paraphrase performance, t-tests for dataset comparisons, and ANOVA for differences across causal types.

\section{Experimental Design}\label{sec:exp}


We specifically select models confirmed to be trained on \emph{The Pile} dataset \cite{Gao2020,Phang2022eleutherai}. 
Since PubMed abstracts used in \citet{Yu2019} are included in \emph{The Pile}, these models necessarily encountered our MCQA sentences during pretraining. As a control, we also include a model not trained on \emph{The Pile}, allowing us to distinguish data exposure effects from other confounders.

In total, we evaluate seven models spanning different architectures, training data, and training approaches. Base pretrained models include pythia variants (1.4B, 7B, 12B parameters) and gpt-j-6b, all trained on \emph{The Pile} without instruction tuning. Instruction-tuned models include dolly-v2 variants (7B, 12B), which use pythia as base models but undergo additional instruction tuning, and qwen-7b-base, an instruction-tuned model not trained on \emph{The Pile}
\footnote{There is also no explicit claim of biomedical literature in the training data, but exposure to PubMed abstracts through other sources cannot be completely ruled out.}.
This selection allows us to isolate the effects of: (i) exposure to training data (\emph{The Pile}), (ii) model scale, and (iii) instruction tuning on causal understanding and uncertainty.

To investigate the link between uncertainty sources and familiar causal patterns in the context of causal understanding, we focus on two complementary tasks.


\textbf{Task 1: Causal Type Classification.} 
Given a sentence $s \in \mathcal{S}$, where $\mathcal{S}$ represents the set of scientific conclusion sentences, the model $M$ must classify the causal relationship into one of four  predefined classes: $\mathcal{Y} =$ \{causal, conditional causal, correlational, no relationship\}. Formally, the model applies a mapping function:
\begin{equation}
f_M: \mathcal{S} \rightarrow \mathcal{Y}\,,
\end{equation}
returning the predicted class $\hat{y}:=f_M(s)$ for the sentence $s$. In particular, we are interested in probabilistic models, returning a probability distribution over the four causal types. In these cases, the model's prediction is:
\begin{equation}
\hat{y} = \arg\max_{y \in \mathcal{Y}} P_M(y|s)
\end{equation}
This directly tests causal understanding with a random baseline of $\frac{1}{|\mathcal{Y}|} = 25\%$.


\textbf{Task 2: Verbatim Memorization Probing.} 
Following \citet{Duarte2024}'s hypothesis that models preferentially select exact text included in their training data (verbatim recall), we test for memorization bias in the context of seen-versus-unseen text. Given a question $q$ derived from an original sentence $s_0 \in \mathcal{X}_{Original \cup Newer}$ and a set of semantically equivalent paraphrases $\mathcal{P} = \{s_0, s_1, \ldots, s_n\}$ where $\text{meaning}(s_i) = \text{meaning}(s_0)$ for all $i$, where $\text{meaning(.)} $ indicates the semantics of the sentence, the model must select the most appropriate answer:
\begin{equation}
\hat{s} = \arg\max_{s_i \in \mathcal{P}} P_M(s_i|q)
\end{equation}
Under the memorization hypothesis, we expect:
\begin{equation}
P_M(s_0|q) > P_M(s_i|q) \quad \forall i \in \{1, \ldots, n\}
\end{equation}
when $s_0$ was seen during training. In contrast, without memorization bias, we expect uniform selection probability: $P_M(s_i|q) \approx \frac{1}{|\mathcal{P}|}$ for all $i$.

To mitigate documented selection biases (e.g., positional bias) in multiple-choice questions, we randomize the order of answer options for each question.  
All models are self-hosted and queried via VLLM \cite{kwon2023efficient} API with temperature $0.0$ for deterministic outputs.
For each prediction, we extract: (i) the selected choice, (ii) the probability distribution over choices, (iii) the corresponding entropy, and (iv) the maximum probability (confidence).

\paragraph{Memorization Assumptions.}
We acknowledge that presence in \emph{The Pile} does not guarantee memorization. Previous work \cite{Carlini2023, Kandpal2023} shows reliable memorization requires 100+ exact repetitions during training. Our design tests whether exposure (seen patterns) — even without guaranteed memorization—provides any advantage for causal reasoning.

\section{Results and Analysis}\label{sec:result}

\subsection{Pretraining-observance Does Not Reduce Uncertainty}

Models show no performance advantage on training data. 
Across all models trained on \emph{The Pile}, accuracy differs by <1.5\% between MCQA and MCQA-newer (Table \ref{tab:summary}). All Pile-trained models with high entropy ($>$1.3) perform near random chance (25\%) but with appropriate uncertainty.
Statistical significance tests confirm these observations (Table~\ref{tab:statistical_tests} and in Appendix  \ref{sec:appendix_significance}). T-tests comparing original versus paraphrase performance yield p-values > 0.05 for all models, with negligible effect sizes (Cohen's d < 0.2). Dataset comparisons (MCQA vs MCQA-newer) show similar results with small effect sizes (|d| < 0.2), indicating no systematic advantage on the data observed in pre-training, suggesting that the presence of an example in the pre-training corpus does not reliably lead to verbatim recall or systematic memorization that could benefit accuracy on the task.

\begin{figure*}[hbt]
    \centering
    \begin{minipage}{0.96\textwidth}
        \centering
        \includegraphics[width=\linewidth]{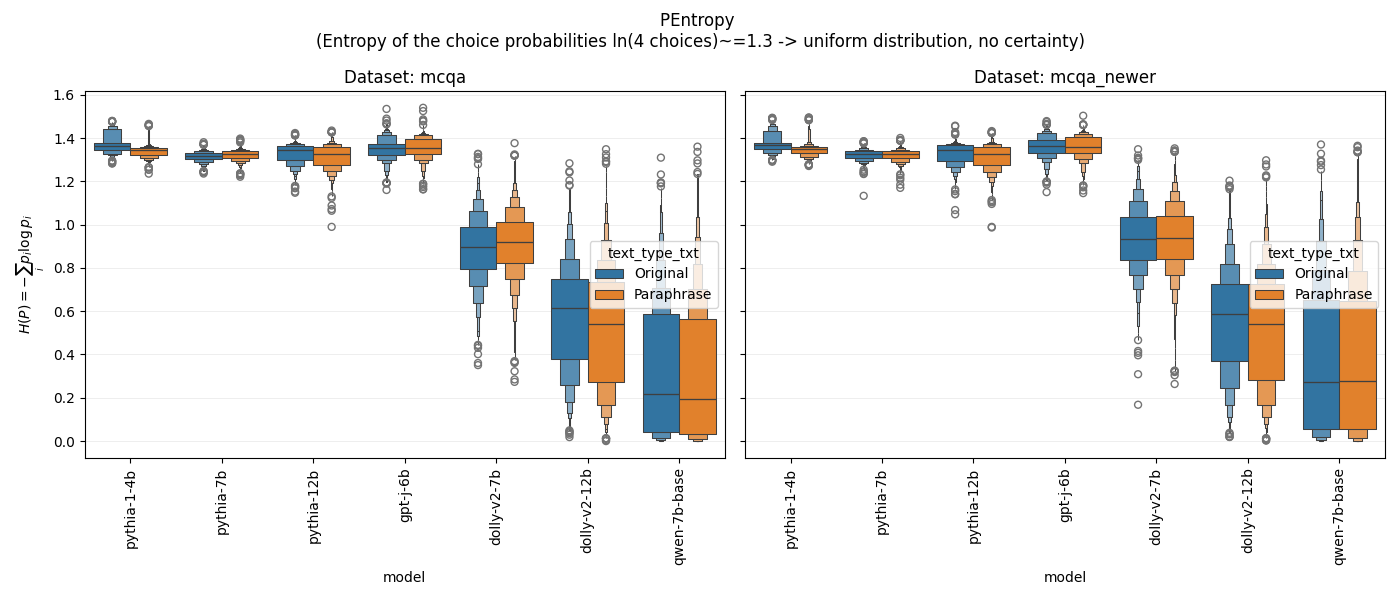}
        \caption{
            \textbf{Entropy distributions across models.}
            Higher values indicate greater uncertainty (max=1.39 for random guessing). Qwen-7b shows low entropy (high confidence) while Pythia/GPT-J show near-maximum entropy.            
        }
        \label{fig:p_entropy2}
    \end{minipage}
\end{figure*}

Figure \ref{fig:p_entropy2} reveals that entropy remains consistently high across both datasets. Pythia and GPT-J models exhibit entropy near maximum (1.3-1.35), indicating near-random guessing regardless of data familiarity. Only qwen-7b-base, not trained on \emph{The Pile}, achieves lower entropy (0.29), suggesting better causal understanding is likely attributable to model architecture rather than mere training data exposure.

To better understand these patterns, we also conducted statistical tests on entropy measures in Table~\ref{tab:statistical_tests} (see also \ref{sec:appendix_significance} in the appendix).
The results reveal a striking divergence: while accuracy remains stable across original and paraphrased sentences, entropy patterns differ significantly (p < 0.001 for pythia and dolly models).
This divergence indicates that models exhibit \textit{different types of uncertainty} for \textit{familiar} versus novel phrasings, even when performing equally poorly. Specifically, pythia models show higher entropy (more uniform distributions) on paraphrases, suggesting they become \textit{more uncertain} when surface forms change. This pattern persists despite no accuracy improvement on original sentences, providing strong evidence against functional memorization.
The MCQA vs MCQA-newer comparisons support this: entropy differences are significant for several models (pythia-7b, gpt-j-6b, dolly models) while accuracy remains constant. 

Models develop different uncertainty profiles for familiar versus unfamiliar datasets without corresponding performance benefits.
Only qwen-7b-base approaches significance for accuracy on original vs paraphrase (p = 0.082), suggesting instruction tuning may introduce subtle biases toward familiar phrasings. However, the effect size remains negligible (d = 0.023).
ANOVA tests on the entropy measures reveal also significant differences across causal types (p < 0.001), confirming that uncertainty patterns reflect task complexity rather than data familiarity (details of the probability and entropy assignments are shown in Appendix \ref{app:prob}. 

\begin{table*}[h]
    \centering
    \small
    \begin{tabular}{lccccccc}
    \toprule
    Model & Accuracy & Entropy & ECE & ACE & \multicolumn{2}{c}{Accuracy} & $\Delta$\\
     & (Overall) & (Mean$\pm$SD) & &  & Original & Paraphrase & (O-P)\\
    \midrule
    pythia-1.4b & 0.248 & 1.34$\pm$0.05 & 0.067 & 0.136 & 0.245 & 0.249 & -0.004\\
    pythia-7b & 0.251 & 1.32$\pm$0.06 & 0.131 & 0.142  & 0.249 & 0.252 & -0.003\\
    pythia-12b & 0.231 & 1.32$\pm$0.07 & 0.149 & 0.151  & 0.231 & 0.231 & 0.000\\
    gpt-j-6b & 0.175 & 1.35$\pm$0.04 & 0.151 & 0.158  & 0.169 & 0.177 & -0.008\\
    dolly-v2-7b & 0.240 & 0.91$\pm$0.19 & 0.363 & 0.239  & 0.240 & 0.240 & 0.000\\
    dolly-v2-12b & 0.212 & 0.53$\pm$0.28 & 0.564 & 0.312  & 0.202 & 0.215 & -0.013\\
    qwen-7b-base & 0.328 & 0.29$\pm$0.35 & 0.493 & 0.275  & 0.339 & 0.326 & 0.013\\
    \bottomrule
    \end{tabular}
    \caption{Summary statistics for causal type classification task. ECE (lower is better). ACE (lower is better). Worst-performing models show best calibration, while better-performing models exhibit overconfidence.}
    
    \label{tab:summary}
\end{table*}


\begin{table}[h]
\centering
\small
\begin{tabular}{lcccc}
\toprule
& \multicolumn{2}{c}{Accuracy} & \multicolumn{2}{c}{Entropy} \\
\cmidrule(lr){2-3} \cmidrule(lr){4-5}
Model & OvP & MvN & OvP & MvN \\
\midrule
    pythia-1.4b & 0.744 & 0.604  & \textbf{<0.001} & 0.573 \\
    pythia-7b & 0.859 & 0.708  & \textbf{<0.001} & \textbf{<0.001} \\
    pythia-12b & 0.823 & 0.506 & \textbf{<0.001} & 0.513 \\
    gpt-j-6b & 0.670 & 0.655  & 0.787 & \textbf{<0.001} \\
    dolly-v2-7b & 0.625 & 0.589 & \textbf{<0.001} & \textbf{<0.001} \\
    dolly-v2-12b & 0.525 & \textbf{0.001} & \textbf{<0.001} & \textbf{0.004} \\
    qwen-7b-base & \textbf{0.082} & 0.672 & 0.123 & \textbf{<0.001} \\
\bottomrule
\end{tabular}
\caption{Statistical significance tests (p-values). OvP: Original vs Paraphrase; MvN: MCQA vs MCQA-Newer. Bold indicates p < 0.05.
Note the divergence between accuracy and entropy tests, revealing that uncertainty patterns differ from performance patterns.}
\label{tab:statistical_tests}
\end{table}

\subsection{Two Distinct Uncertainty Profiles}

\begin{figure*}[hbt]
    \centering
    \begin{minipage}{0.96\textwidth}
        \centering
        \includegraphics[width=\linewidth]{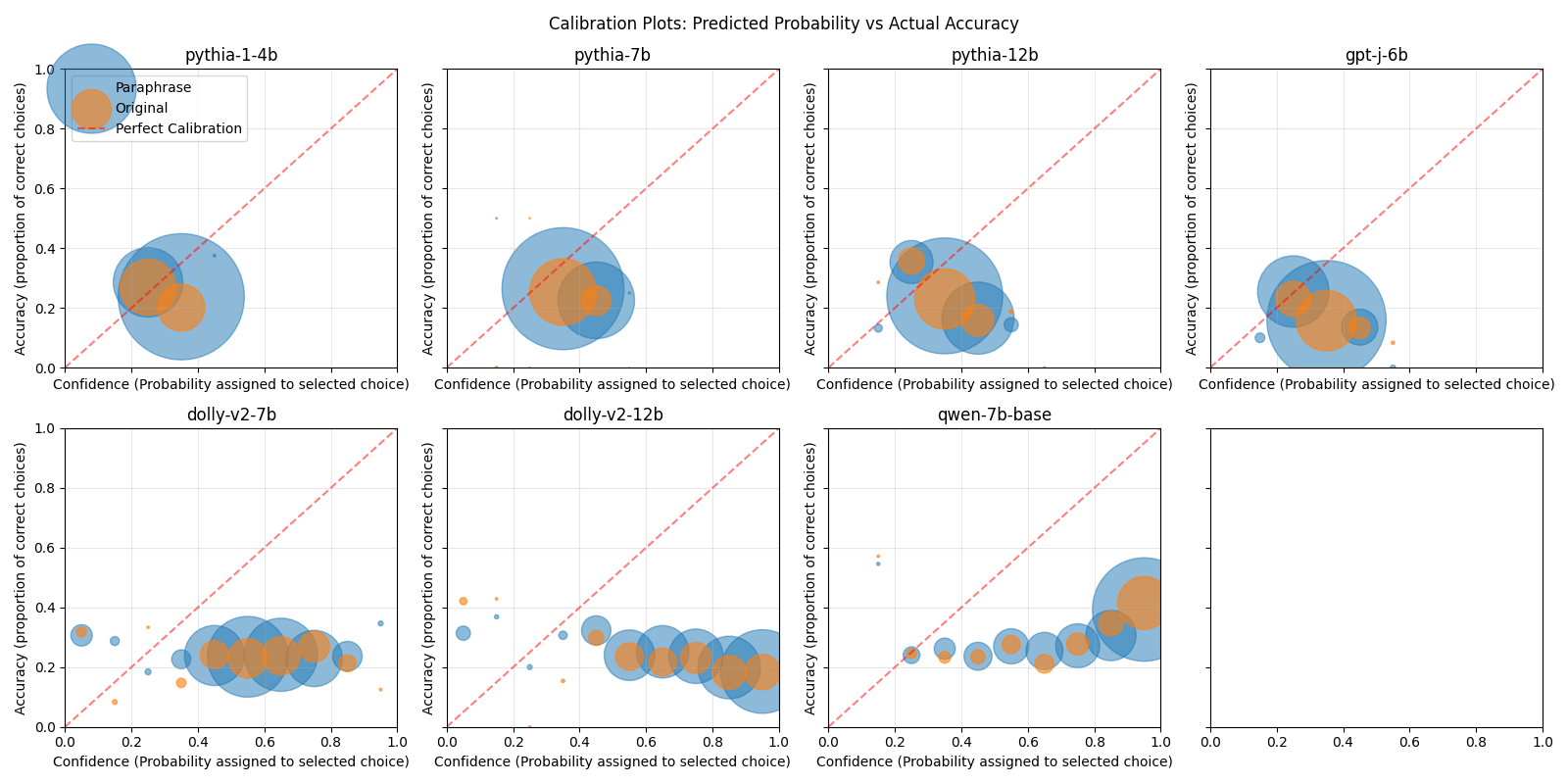}
        \caption{
            \textbf{Calibration plots, confidence vs accuracy.}
            Perfect calibration follows the diagonal. Base models (Pythia/GPT-J) show good calibration despite poor accuracy. Instruction-tuned models (Qwen/Dolly) show overconfidence.
        }
        \label{fig:calibration}
    \end{minipage}
\end{figure*}

Figure~\ref{fig:calibration} reveals two distinct calibration patterns: base models maintain appropriate uncertainty despite 25\% accuracy, while instruction-tuned models exhibit overconfidence.

The distribution of prediction confidence (see Appendix \ref{app:overconfidence}) 
reveals two distinct patterns. Pile-trained models (pythia, gpt-j) consistently assign low confidence to their predictions, with probability distributions peaked around 0.30-0.35—appropriately uncertain given their near-random accuracy. These models maintain ECE < 0.16, indicating well-calibrated uncertainty.

In contrast, qwen-7b-base exhibits overconfidence, with most predictions assigned >95\% probability despite achieving only 32.8\% accuracy. 
This confidence-accuracy gap yields ECE=0.49 versus 0.13 for base models—a 3.8x increase in calibration error.
Dolly models, with accuracies between 21\% and 23\%, show bimodal confidence distributions but with very different calibration error patterns depending on the model size, i.e., the 7B version has an ECE of 0.36 and ACE of 0.23, whereas 12B although less accurate than the 7B,  presents the highest calibration error of all models (ECE=0.56, ACE=0.31).

These observations suggest that in causal understanding tasks, partial competence breeds false confidence. Models performing near random maintain appropriate low confidence, while the best-performing model develops overconfidence—an especially concerning trait in applications that demand reliable causal inference.

\subsection{Instruction Tuning Creates Overconfidence}

Our model selection reveals an important factor: instruction tuning fundamentally alters uncertainty behavior in causal understanding. Base pretrained models (pythia variants, gpt-j-6b) exhibit high entropy ($\approx$1.35) with appropriately low confidence (30-35\%), yielding good calibration despite poor performance.

However, instruction-tuned models show different patterns. Dolly models—fine-tuned from pythia bases on instruction-following data—develop moderate confidence (40-60\%) without corresponding accuracy improvements. Most markedly, qwen-7b-base exhibits overconfidence (>95\%) while achieving only marginally better accuracy (32.8\%).

This divergence offers valuable insights into dolly models: 
identical pretrained weights (pythia) produce different uncertainty profiles after instruction tuning. Compare pythia-7b (ECE=0.13, entropy=1.32) with dolly-v2-7b (ECE=0.36, entropy=0.92)—instruction tuning does not even half the entropy while almost tripling the calibration error.

These results suggest instruction tuning teaches models to be confident in their responses, even when this confidence is unjustified. While this may improve user experience in conversational settings, it creates problematic overconfidence in domains requiring accurate uncertainty quantification, such as causal-language related tasks.

\subsection{Uncertainty Varies by Causal Complexity}

Uncertainty patterns differ across causal relationship types. Conditional causal statements induce the highest entropy across all models, while direct causal relationships show moderately lower uncertainty. Correlational statements, despite being non-causal, often receive more confident predictions than conditional causal ones.

ANOVA confirms significant differences in entropy across causal types (p < 0.001 for all models). This pattern persists across both original and paraphrase sentences, indicating that uncertainty arises from conceptual difficulty rather than surface-level confusion.

\begin{table}[h]
    \centering
    \small
    \begin{tabular}{lcc}
    \toprule
    Causal Type & Mean Entropy & Accuracy\\
    \midrule
    Direct Causal & 1.15$\pm$0.42 & 0.31\\
    Conditional Causal & 1.28$\pm$0.29 & 0.19\\
    Correlational & 1.09$\pm$0.45 & 0.26\\
    No Relationship & 1.12$\pm$0.43 & 0.22\\
    \bottomrule
    \end{tabular}
    \caption{Performance breakdown by causal relationship type (averaged across all models). Conditional causal statements show highest entropy and lowest accuracy.}
    \label{tab:causal_types}
\end{table}

Table~\ref{tab:causal_types} quantifies this pattern: conditional causal statements exhibit 11\% higher entropy than direct causal statements, approaching maximum entropy. This suggests that models struggle particularly with nuanced causal identifications involving conditions or moderating factors. Detailed analysis can be found in Appendix \ref{app:uncertainty}. 

\subsection{Identifying Inherently Ambiguous Questions}
We analyzed the results based on the intuition that if all paraphrases of a sentence get similar wrong predictions, it might indicate inherent ambiguity (aleatoric uncertainty).
    
\begin{figure}[h!bt]
    \centering
    \includegraphics[width=0.45\textwidth]{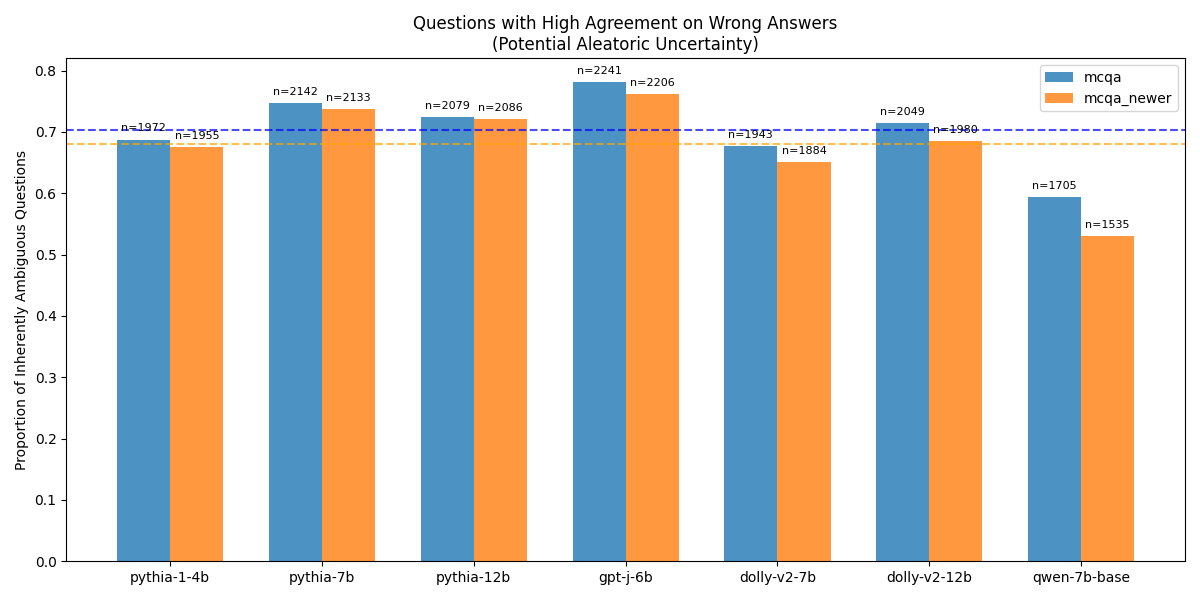}
    \caption{
        \textbf{Ambiguous questions}
        Proportion of questions with consistent wrong answers across paraphrases, indicating inherent ambiguity.
    }
    \label{fig:ambiguous}
\end{figure}

This paraphrase consistency analysis reveals a subset of questions where all models consistently select the same incorrect answer across paraphrases (i.e., consistency > 0.7 and accuracy < 0.3). These represent 60-75\% of questions depending on the model (Figure \ref{fig:ambiguous}). High consistency on wrong answers suggests inherent ambiguity in the task rather than model-specific confusion.

A manual analysis of misclassified instances reveals recurring linguistic patterns, including the use of hedging expressions (e.g., “may influence,” “suggests association”) and complex multi-clause constructions. These features are associated with misclassification across model families.

\subsection{No Evidence on Verbatim Recalls with Pre-training Exposures}


The verbatim memorization probing (task 2) provides complementary evidence. When presented with questions about causal relationships and asked to choose between original sentences and paraphrases (all semantically correct), models show no preference for the original form. Selection rates for original sentences average 24.8\% (95\% CI: [24.2\%, 25.4\%]) across all Pile-trained models, statistically indistinguishable from the 25\% random baseline (chi-square test, p > 0.05).

Entropy analysis of this task reveals similar patterns: models exhibit high uncertainty (mean entropy 1.31) regardless of whether choosing between already seen or potentially novel phrasings. This task design, which controls for correctness while varying surface form, shows that models do not leverage already seen patterns even when given the opportunity.

This null result holds even when examining only high-frequency scientific phrases likely repeated many times in training data. Models treat original and paraphrase sentences identically, 
confirming that surface-level causal patterns exposed during pretraining may not contribute to causal understanding (details of probabilities and entropy analysis is in Appendix \ref{app:task2}). 

\begin{figure*}[hbt]
    \centering
    \begin{minipage}{0.96\textwidth}
        \centering
        \includegraphics[width=\linewidth]{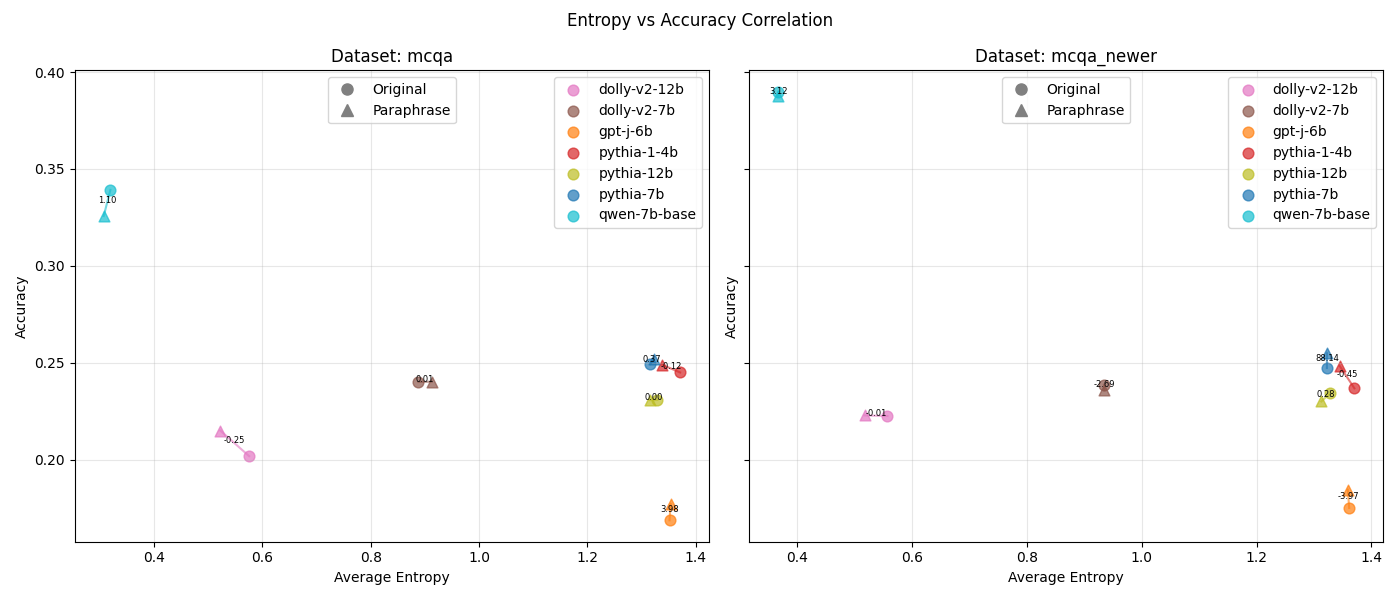}
        \caption{
            \textbf{Entropy vs Accuracy.}
            Ideal models appear top-left (accurate and confident). Most models cluster bottom-right (inaccurate and uncertain).
        }        
        \label{fig:entropyvsaccuracy}
    \end{minipage}
\end{figure*}

\subsection{Consistency Across Semantic Paraphrases}
To further test verbatim recall, we analyze prediction consistency across paraphrases of the same sentence. If models rely on "seen/recalled" patterns, we expect higher consistency for original sentences versus their paraphrases. 

For each sentence with multiple probes, we computed Pearson correlation between choice probability vectors to assess confidence consistency, Spearman rank correlation to evaluate preference consistency by comparing ranked orders, and KL divergence to quantify overall distributional shifts between probabilistic outputs (see Appendix \ref{app:consistency} for details).

Only pretrained models exhibit moderate KL divergence (KL < 0.15), suggesting relatively stable probabilistic behavior across input variations.
However, larger and instruction-tuned models such as qwen-7b-base and dolly models exhibit higher KL divergence (>0.5 for qwen and >1.0 for dolly models), indicating greater sensitivity to paraphrasing and less consistency in their output distributions, i.e., unstable predictions across semantically equivalent inputs. This pattern holds across both the MCQA and MCQA-newer datasets.

Correlation analysis of probability vectors and ranked order reveals weak consistency (mean Pearson and Spearman $\rho$ < 0.2) across all models but qwen-7b-base (Pearson $\rho$ > 0.71 and Spearman $\rho$ > 0.73).

Critically, consistency patterns do not differ between MCQA (training data) and MCQA-newer (unseen data). The absence of improved consistency on training data provides evidence against memorization-based reasoning. Models treat semantically equivalent sentences as unrelated inputs, confirming they lack robust causal understanding rather than merely forgetting training examples.

\section{Discussion}\label{sec:discuss}
Our experimental findings show that uncertainty in causal tasks stems from absent causal understanding rather than insufficient data exposure (Figure \ref{fig:entropyvsaccuracy}).
Three key insights emerge:
\textbf{First}, the independence of performance from training data exposure suggests that simply scaling datasets will not resolve causal understanding deficits. Models require architectural innovations or training objectives that explicitly target causal inference.
\textbf{Second}, 
overconfidence from instruction tuning poses deployment risks. The shift from calibrated uncertainty (base models: ECE=0.13) to overconfidence (instruction-tuned: ECE=0.49) indicates fine-tuning teaches models to suppress appropriate uncertainty.
\textbf{Third}, the particular difficulty with conditional causal relationships indicates that models lack compositional reasoning about causality. While they may recognize simple cause-and-effect patterns, they fail when conditions, moderators, or exceptions are introduced.

These results suggest that causal understanding in LLMs requires fundamental advances beyond current pretraining paradigms. Memorization, even at scale, cannot substitute for genuine causal knowledge.

\section{Conclusion}\label{sec:concl}
In this work, we focused on analyzing the critical limitations in the causal understanding abilities of large language models (LLMs). Through a controlled evaluation combining causal classification and verbatim memorization probing, we demonstrate that exposure to causal content during pretraining does not guarantee accurate recall or improved causal understanding. Our analysis, leveraging multiple uncertainty metrics—including entropy, consistency, calibration, and accuracy—reveals that uncertainty in causal tasks stems primarily from deficits in causal understanding rather than limitations in memorization. Addressing these limitations will require a shift beyond current pretraining paradigms—toward models that explicitly encode and reason over causal structures, and that are capable of expressing calibrated uncertainty when faced with ambiguity or unseen conditions.

\section*{Limitations}
Our study has three key limitations. First, we cannot determine whether the models truly failed to acquire causal patterns during training, or whether they learned them but are unable to apply or recall them during inference. Structured prompting, Causal probing with small datasets, pre-trained data inspections through sampling, probing representations, etc., can be possible approaches to tackle this problem. Second, presence in \emph{The Pile} does not guarantee memorization—research shows reliable memorization requires 100+ repetitions \cite{Kandpal2023}. Our results demonstrate that even exposure without guaranteed memorization provides no benefit for causal reasoning. Third, our binary classification of "seen" versus "unseen" may oversimplify the memorization spectrum. Future work should examine the relationship between repetition frequency and causal understanding. 

\section*{Acknowledgments}
This work has been supported by UBS Switzerland AG and its affiliates.

\bibliography{emnlp2025_uncertainty}

\begin{thebibliography}{29}
\providecommand{\natexlab}[1]{#1}

\bibitem[{Ashwani et~al.(2024)Ashwani, Hegde, Mannuru, Jindal, Sengar, Kathala, Banga, Jain, and Chadha}]{Ashwani2024}
Swagata Ashwani, Kshiteesh Hegde, Nishith~Reddy Mannuru, Mayank Jindal, Dushyant~Singh Sengar, Krishna Chaitanya~Rao Kathala, Dishant Banga, Vinija Jain, and Aman Chadha. 2024.
\newblock \href {https://doi.org/10.48550/arXiv.2402.18139} {Cause and {{Effect}}: {{Can Large Language Models Truly Understand Causality}}?}
\newblock \emph{Preprint}, arXiv:2402.18139.

\bibitem[{Carlini et~al.(2023)Carlini, Ippolito, Jagielski, Lee, Tramer, and Zhang}]{Carlini2023}
Nicholas Carlini, Daphne Ippolito, Matthew Jagielski, Katherine Lee, Florian Tramer, and Chiyuan Zhang. 2023.
\newblock \href {https://doi.org/10.48550/arXiv.2202.07646} {Quantifying {{Memorization Across Neural Language Models}}}.
\newblock \emph{Preprint}, arXiv:2202.07646.

\bibitem[{Carlini et~al.(2021)Carlini, Tramer, Wallace, Jagielski, Herbert-Voss, Lee, Roberts, Brown, Song, Erlingsson, Oprea, and Raffel}]{Carlini2021}
Nicholas Carlini, Florian Tramer, Eric Wallace, Matthew Jagielski, Ariel Herbert-Voss, Katherine Lee, Adam Roberts, Tom Brown, Dawn Song, Ulfar Erlingsson, Alina Oprea, and Colin Raffel. 2021.
\newblock \href {https://doi.org/10.48550/arXiv.2012.07805} {Extracting {{Training Data}} from {{Large Language Models}}}.
\newblock \emph{Preprint}, arXiv:2012.07805.

\bibitem[{Cui et~al.(2025)Cui, Mouchel, and Faltings}]{Cui2025}
Shaobo Cui, Luca Mouchel, and Boi Faltings. 2025.
\newblock \href {https://doi.org/10.18653/v1/2025.acl-long.396} {Uncertainty in {{Causality}}: {{A New Frontier}}}.
\newblock In \emph{Proceedings of the 63rd {{Annual Meeting}} of the {{Association}} for {{Computational Linguistics}} ({{Volume}} 1: {{Long Papers}})}, pages 8022--8044. Association for Computational Linguistics.

\bibitem[{Duarte et~al.(2024)Duarte, Zhao, Oliveira, and Li}]{Duarte2024}
André~V. Duarte, Xuandong Zhao, Arlindo~L. Oliveira, and Lei Li. 2024.
\newblock \href {https://doi.org/10.48550/arXiv.2402.09910} {{{DE-COP}}: {{Detecting Copyrighted Content}} in {{Language Models Training Data}}}.
\newblock \emph{Preprint}, arXiv:2402.09910.

\bibitem[{Feng et~al.(2024)Feng, Qu, Tandon, Li, Kang, and Haffari}]{Feng2024b}
Tao Feng, Lizhen Qu, Niket Tandon, Zhuang Li, Xiaoxi Kang, and Gholamreza Haffari. 2024.
\newblock \href {https://doi.org/10.48550/arXiv.2407.19638} {From {{Pre-training Corpora}} to {{Large Language Models}}: {{What Factors Influence LLM Performance}} in {{Causal Discovery Tasks}}?}
\newblock \emph{Preprint}, arXiv:2407.19638.

\bibitem[{Gao et~al.(2020)Gao, Biderman, Black, Golding, Hoppe, Foster, Phang, He, Thite, Nabeshima, Presser, and Leahy}]{Gao2020}
Leo Gao, Stella Biderman, Sid Black, Laurence Golding, Travis Hoppe, Charles Foster, Jason Phang, Horace He, Anish Thite, Noa Nabeshima, Shawn Presser, and Connor Leahy. 2020.
\newblock \href {https://doi.org/10.48550/arXiv.2101.00027} {The {{Pile}}: {{An 800GB Dataset}} of {{Diverse Text}} for {{Language Modeling}}}.
\newblock \emph{Preprint}, arXiv:2101.00027.

\bibitem[{Giulianelli et~al.(2023)Giulianelli, Baan, Aziz, Fernández, and Plank}]{Giulianelli2023}
Mario Giulianelli, Joris Baan, Wilker Aziz, Raquel Fernández, and Barbara Plank. 2023.
\newblock \href {https://doi.org/10.18653/v1/2023.emnlp-main.887} {What {{Comes Next}}? {{Evaluating Uncertainty}} in {{Neural Text Generators Against Human Production Variability}}}.
\newblock In \emph{Proceedings of the 2023 {{Conference}} on {{Empirical Methods}} in {{Natural Language Processing}}}, pages 14349--14371. Association for Computational Linguistics.

\bibitem[{Guo et~al.(2017)Guo, Pleiss, Sun, and Weinberger}]{Guo2017}
Chuan Guo, Geoff Pleiss, Yu~Sun, and Kilian~Q. Weinberger. 2017.
\newblock \href {https://proceedings.mlr.press/v70/guo17a.html} {On {{Calibration}} of {{Modern Neural Networks}}}.
\newblock In \emph{Proceedings of the 34th {{International Conference}} on {{Machine Learning}}}, pages 1321--1330. PMLR.

\bibitem[{He et~al.(2025)He, Tech, Yu, Li, Yang, Tech, Chen, Tech, Li, Zhang, Tech, Lei, Tech, Zhang, Beigi, Ding, Xiao, Huang, Chen, Jin, Tech, Lu, and Tech}]{He}
Jianfeng He, Virginia Tech, Linlin Yu, Changbin Li, Runing Yang, Virginia Tech, Fanglan Chen, Virginia Tech, Kangshuo Li, Min Zhang, Virginia Tech, Shuo Lei, Virginia Tech, Xuchao Zhang, Mohammad Beigi, Kaize Ding, Bei Xiao, Lifu Huang, Feng Chen, and 4 others. 2025.
\newblock Survey of {{Uncertainty Estimation}} in {{LLMs}} - {{Sources}}, {{Methods}}, {{Applications}}, and {{Challenge}}.
\newblock 1(1).

\bibitem[{Hüllermeier and Waegeman(2021)}]{Hullermeier2021}
Eyke Hüllermeier and Willem Waegeman. 2021.
\newblock \href {https://doi.org/10.1007/s10994-021-05946-3} {Aleatoric and {{Epistemic Uncertainty}} in {{Machine Learning}}: {{An Introduction}} to {{Concepts}} and {{Methods}}}.
\newblock 110(3):457--506.

\bibitem[{Joshi et~al.(2024)Joshi, Saparov, Wang, and He}]{Joshi2024a}
Nitish Joshi, Abulhair Saparov, Yixin Wang, and He~He. 2024.
\newblock \href {https://doi.org/10.18653/v1/2024.emnlp-main.590} {{{LLMs Are Prone}} to {{Fallacies}} in {{Causal Inference}}}.
\newblock In \emph{Proceedings of the 2024 {{Conference}} on {{Empirical Methods}} in {{Natural Language Processing}}}, pages 10553--10569. Association for Computational Linguistics.

\bibitem[{Kandpal et~al.(2023)Kandpal, Deng, Roberts, Wallace, and Raffel}]{Kandpal2023}
Nikhil Kandpal, Haikang Deng, Adam Roberts, Eric Wallace, and Colin Raffel. 2023.
\newblock \href {https://doi.org/10.48550/arXiv.2211.08411} {Large {{Language Models Struggle}} to {{Learn Long-Tail Knowledge}}}.
\newblock \emph{Preprint}, arXiv:2211.08411.

\bibitem[{Kanjirangat et~al.(2024)Kanjirangat, Antonucci, and Zaffalon}]{Kanjirangat2024}
Vani Kanjirangat, Alessandro Antonucci, and Marco Zaffalon. 2024.
\newblock \href {https://openreview.net/forum?id=QZSfMkdv6y} {On the {{Limitations}} of {{Zero-Shot Classification}} of {{Causal Relations}} by {{LLMs}}}.

\bibitem[{Kirchhof et~al.(2025)Kirchhof, Füger, Goliński, Dhekane, Blaas, and Williamson}]{Kirchhof2025a}
Michael Kirchhof, Luca Füger, Adam Goliński, Eeshan~Gunesh Dhekane, Arno Blaas, and Sinead Williamson. 2025.
\newblock \href {https://doi.org/10.48550/arXiv.2505.20295} {Self-reflective {{Uncertainties}}: {{Do LLMs Know Their Internal Answer Distribution}}?}
\newblock \emph{Preprint}, arXiv:2505.20295.

\bibitem[{Kwon et~al.(2023)Kwon, Li, Zhuang, Sheng, Zheng, Yu, Gonzalez, Zhang, and Stoica}]{kwon2023efficient}
Woosuk Kwon, Zhuohan Li, Siyuan Zhuang, Ying Sheng, Lianmin Zheng, Cody~Hao Yu, Joseph~E. Gonzalez, Hao Zhang, and Ion Stoica. 2023.
\newblock Efficient memory management for large language model serving with {{PagedAttention}}.
\newblock In \emph{Proceedings of the {{ACM SIGOPS}} 29th Symposium on Operating Systems Principles}.

\bibitem[{Li et~al.(2024)Li, Zhao, and Wen}]{Li2024f}
Bo~Li, Qinghua Zhao, and Lijie Wen. 2024.
\newblock \href {https://doi.org/10.48550/arXiv.2403.00510} {{{ROME}}: {{Memorization Insights}} from {{Text}}, {{Logits}} and {{Representation}}}.
\newblock \emph{Preprint}, arXiv:2403.00510.

\bibitem[{Liu et~al.(2025)Liu, Chen, Da, Chen, Lin, and Wei}]{Liu2025d}
Xiaoou Liu, Tiejin Chen, Longchao Da, Chacha Chen, Zhen Lin, and Hua Wei. 2025.
\newblock \href {https://doi.org/10.48550/arXiv.2503.15850} {Uncertainty {{Quantification}} and {{Confidence Calibration}} in {{Large Language Models}}: {{A Survey}}}.
\newblock \emph{Preprint}, arXiv:2503.15850.

\bibitem[{Nixon et~al.(2019{\natexlab{a}})Nixon, Dusenberry, Zhang, Jerfel, and Tran}]{nixon2019measuring}
Jeremy Nixon, Michael~W Dusenberry, Linchuan Zhang, Ghassen Jerfel, and Dustin Tran. 2019{\natexlab{a}}.
\newblock Measuring calibration in deep learning.
\newblock In \emph{CVPR workshops}, volume~2.

\bibitem[{Nixon et~al.(2019{\natexlab{b}})Nixon, Dusenberry, Zhang, Jerfel, and Tran}]{Nixon2019}
Jeremy Nixon, Michael~W. Dusenberry, Linchuan Zhang, Ghassen Jerfel, and Dustin Tran. 2019{\natexlab{b}}.
\newblock \href {https://openaccess.thecvf.com/content_CVPRW_2019/html/Uncertainty_and_Robustness_in_Deep_Visual_Learning/Nixon_Measuring_Calibration_in_Deep_Learning_CVPRW_2019_paper.html} {Measuring calibration in deep learning}.
\newblock In \emph{Proceedings of the IEEE/CVF Conference on Computer Vision and Pattern Recognition Workshops}, volume~2, pages 38--41. IEEE / Computer Vision Foundation.

\bibitem[{Phang et~al.(2022)Phang, Bradley, Gao, Castricato, and Biderman}]{Phang2022eleutherai}
Jason Phang, Herbie Bradley, Leo Gao, Louis Castricato, and Stella Biderman. 2022.
\newblock \href {https://doi.org/10.48550/arXiv.2210.06413} {Eleutherai: Going beyond “open science” to “science in the open”}.
\newblock In \emph{Workshop on Broadening Research Collaborations in ML, NeurIPS 2022}. NeurIPS Workshop.

\bibitem[{Posocco and Bonnefoy(2021)}]{posocco2021estimating}
Nicolas Posocco and Antoine Bonnefoy. 2021.
\newblock Estimating expected calibration errors.
\newblock In \emph{International conference on artificial neural networks}, pages 139--150. Springer.

\bibitem[{Shannon(1948)}]{shannon1948mathematical}
Claude~E Shannon. 1948.
\newblock A mathematical theory of communication.
\newblock \emph{The Bell system technical journal}, 27(3):379--423.

\bibitem[{Shorinwa et~al.(2025)Shorinwa, Mei, Lidard, Ren, and Majumdar}]{Shorinwa2025}
Ola Shorinwa, Zhiting Mei, Justin Lidard, Allen~Z. Ren, and Anirudha Majumdar. 2025.
\newblock \href {https://doi.org/10.48550/arXiv.2412.05563} {A {{Survey}} on {{Uncertainty Quantification}} of {{Large Language Models}}: {{Taxonomy}}, {{Open Research Challenges}}, and {{Future Directions}}}.
\newblock \emph{Preprint}, arXiv:2412.05563.

\bibitem[{Tirumala et~al.(2022)Tirumala, Markosyan, Zettlemoyer, and Aghajanyan}]{Tirumala2022}
Kushal Tirumala, Aram~H. Markosyan, Luke Zettlemoyer, and Armen Aghajanyan. 2022.
\newblock \href {https://doi.org/10.48550/arXiv.2205.10770} {Memorization {{Without Overfitting}}: {{Analyzing}} the {{Training Dynamics}} of {{Large Language Models}}}.
\newblock \emph{Preprint}, arXiv:2205.10770.

\bibitem[{Wang et~al.(2024)Wang, Duan, Cheng, Zhang, Wang, Shi, Xu, Shen, and Zhu}]{Wang2024r}
Zhiyuan Wang, Jinhao Duan, Lu~Cheng, Yue Zhang, Qingni Wang, Xiaoshuang Shi, Kaidi Xu, Heng~Tao Shen, and Xiaofeng Zhu. 2024.
\newblock \href {https://doi.org/10.18653/v1/2024.findings-emnlp.404} {{{ConU}}: {{Conformal Uncertainty}} in {{Large Language Models}} with {{Correctness Coverage Guarantees}}}.
\newblock In \emph{Findings of the {{Association}} for {{Computational Linguistics}}: {{EMNLP}} 2024}, pages 6886--6898. Association for Computational Linguistics.

\bibitem[{Yadkori et~al.(2024)Yadkori, Kuzborskij, György, and Szepesvári}]{Yadkori2024}
Yasin~Abbasi Yadkori, Ilja Kuzborskij, András György, and Csaba Szepesvári. 2024.
\newblock \href {https://doi.org/10.48550/arXiv.2406.02543} {To {{Believe}} or {{Not}} to {{Believe Your LLM}}}.
\newblock \emph{Preprint}, arXiv:2406.02543.

\bibitem[{Yu et~al.(2019)Yu, Li, and Wang}]{Yu2019}
Bei Yu, Yingya Li, and Jun Wang. 2019.
\newblock \href {https://doi.org/10.18653/v1/D19-1473} {Detecting causal language use in science findings}.
\newblock In \emph{Proceedings of the 2019 Conference on Empirical Methods in Natural Language Processing and the 9th International Joint Conference on Natural Language Processing (EMNLP-IJCNLP)}, pages 4664--4674. Association for Computational Linguistics.

\bibitem[{Zhang et~al.(2023)Zhang, Ippolito, Lee, Jagielski, Tramèr, and Carlini}]{Zhang2023i}
Chiyuan Zhang, Daphne Ippolito, Katherine Lee, Matthew Jagielski, Florian Tramèr, and Nicholas Carlini. 2023.
\newblock \href {https://doi.org/10.48550/arXiv.2112.12938} {Counterfactual {{Memorization}} in {{Neural Language Models}}}.
\newblock \emph{Preprint}, arXiv:2112.12938.

\end{thebibliography}

\clearpage
\appendix
\appendix
\section{Dataset \& Prompts}\label{app:data}
The examples from the constructed dataset are shown in Figures \ref{fig:examples1} and \ref{fig:examples2}.
\floatstyle{boxed}
\restylefloat{figure}

\floatstyle{boxed}
\restylefloat{figure}
\begin{figure*}[htp!]
\centering
\tiny
\begin{verbatim}
{"qa_idx": 0, "context": "However, the small sample size in this study limits its generalizability to diverse populations, 
so we call for future research that explores SSL-powered personalization at a larger scale.",

"text": "However, the small sample size in this study limits its generalizability to diverse populations, 
so we call for future research that explores SSL-powered personalization at a larger scale.", 
"text_type": 0, "causal_class_label": 0, 

"choices": [{"label": 1, "text": "Direct Causal", "description": "The statement explicitly states that one variable directly causes changes in another."}, 
{"label": 3, "text": "Correlational", "description": "The statement describes an association between variables, but no causation is 
explicitly stated."}, 
{"label": 2, "text": "Conditional Causal", "description": "The statement suggests causation but includes uncertainty through hedging words or
modal expressions."}, 
{"label": 0, "text": "No Relationship", "description": "No correlation or causation relationship is mentioned."}, 
{"label": 4, "text": "Other", "description": ""}]

\end{verbatim}
\caption{Examples from the constructed data (Task 1) - Casual Type Classification}\label{fig:examples1}
\end{figure*}

\begin{figure*}[htp!]
\centering
\tiny
\begin{verbatim}
Example non-causal: {"context": "Faster aspart and IAsp were confirmed noninferior in a basal-bolus regimen regarding change from baseline in HbA1c.", 

"question": "What was the outcome of comparing faster aspart and IAsp in terms of their effect on HbA1c levels in a basal-bolus regimen?  ", 

"choices": [{"text": "Faster aspart and IAsp were shown to be noninferior in a basal-bolus regimen with respect to the change in HbA1c from the starting point.  ",
"type": 1}, 
{"text": "Faster aspart and IAsp were confirmed noninferior in a basal-bolus regimen regarding change from baseline in HbA1c.", "type": 0},
{"text": "Faster aspart and IAsp were validated as noninferior in a basal-bolus treatment concerning the change in HbA1c from baseline.  ", "type": 1}, 
{"text": "Faster aspart and IAsp were not confirmed noninferior in a basal-bolus regimen regarding change from baseline in HbA1c.", "type": 2}, 
{"text": "I don't know", "type": 3}], 
"true_sent_type": 0, "causal_class_label": 0}


Example causal: {"context": "Vildagliptin effectively improved glucose level with a significantly greater reduction in glycemic variability and hypoglycemia 
than glimepiride in patients with T2DM ongoing metformin therapy.", 
"question": "What was the effect of vildagliptin compared to glimepiride on glucose levels in patients with T2DM?  ", 
"choices": [{"text": "Vildagliptin significantly enhanced glucose levels, showing a much larger decrease in glycemic variability
and hypoglycemia compared to glimepiride in patients with T2DM who were already on metformin treatment.  ", "type": 1}, 
{"text": "Vildagliptin effectively improved glucose level with a significantly greater reduction in glycemic variability 
and hypoglycemia than glimepiride in patients with T2DM ongoing metformin therapy.", "type": 0}, 
{"text": "Vildagliptin did not effectively improve glucose level with a significantly greater reduction in glycemic variability 
and hypoglycemia than glimepiride in patients with T2DM ongoing metformin therapy.", "type": 2}, 
{"text": "I don't know", "type": 3}, 
{"text": "In patients with T2DM receiving ongoing metformin therapy, vildagliptin led to a notable improvement in glucose levels, 
with a significantly greater reduction in glycemic variability 
and hypoglycemia than glimepiride.  ", "type": 1}], 
"true_sent_type": 0, "causal_class_label": 1,}

\end{verbatim}
\caption{Examples from the constructed data (Task 2) - Example with the configuration: original (type 0), two paraphrases (type 1), one negation (type 2) and I don't know (type 3), with and without causal labels}\label{fig:examples2}
\end{figure*}

\paragraph{Dataset construction prompt template:} 

\emph{''Paraphrase the following sentence while preserving its exact meaning, especially the causal relationship. Change the wording and structure but keep the scientific accuracy: [sentence]''}

\section{Task 1: Causal Type Classification.}\label{app:task1}

\subsection{Probabilities analysis}\label{app:prob}
\begin{figure}[hbt]
    \centering
        \centering
        \includegraphics[width=1.0\linewidth]{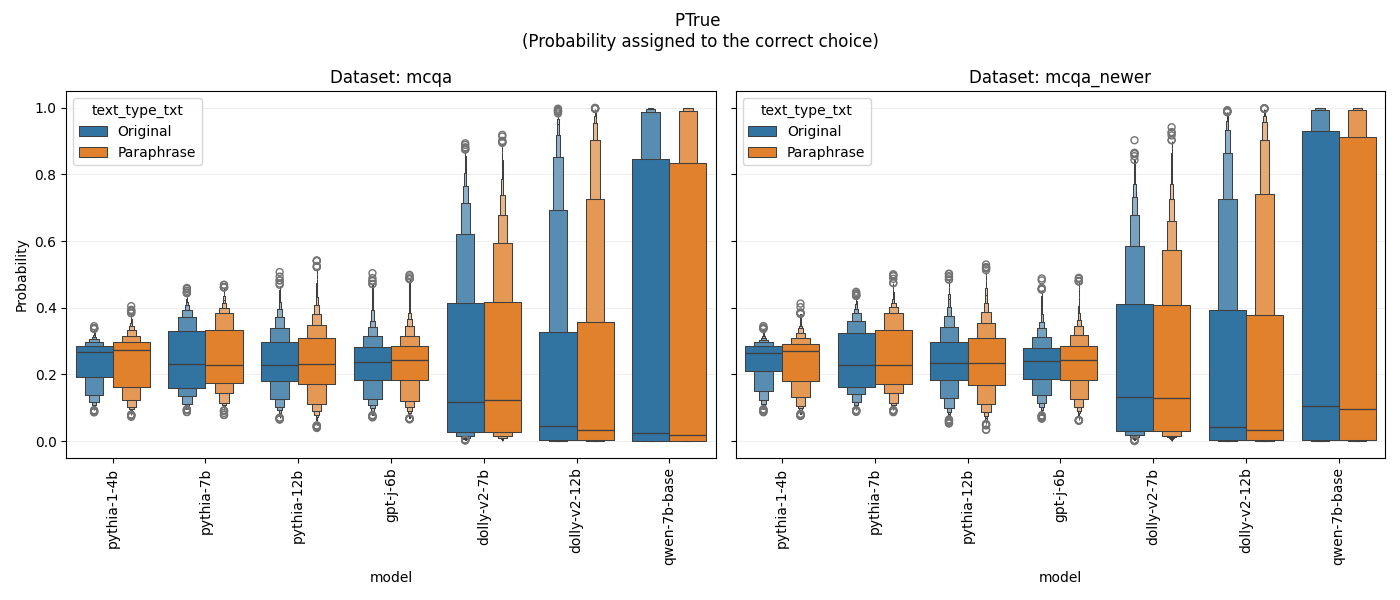}
        \caption{
            \textbf{Probabilities assigned to the correct choice.}
            Box plots showing the distribution of probabilities assigned to correct answers by different models for original questions and paraphrases. Results are shown for \textit{mcqa} (left) and \textit{mcqa\_newer} (right) datasets. Higher probabilities indicate greater model confidence in correct predictions.            
        }
        \label{fig:p_true}
\end{figure}

\begin{figure}[hbt]
    \centering
        \centering
        \includegraphics[width=1.0\linewidth]{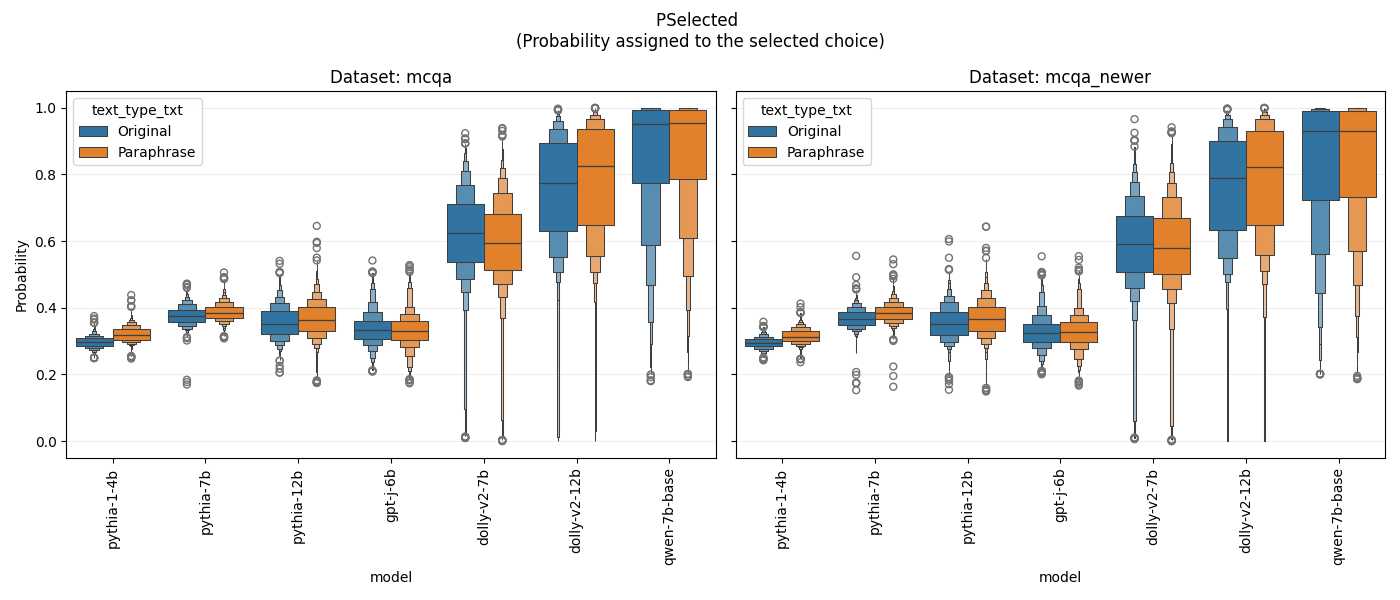}
        \caption{
            \textbf{Probabilities assigned to the selected choice.}
            Box plots showing the distribution of probabilities assigned to selected answers by different models for original questions and paraphrases. Results are shown for \textit{mcqa} (left) and \textit{mcqa\_newer} (right) datasets. Higher probabilities indicate greater model confidence in selected predictions.            
        }
        \label{fig:p_selected}
\end{figure}

\begin{figure}[hbt]
    \centering
        \centering
        \includegraphics[width=1.0\linewidth]{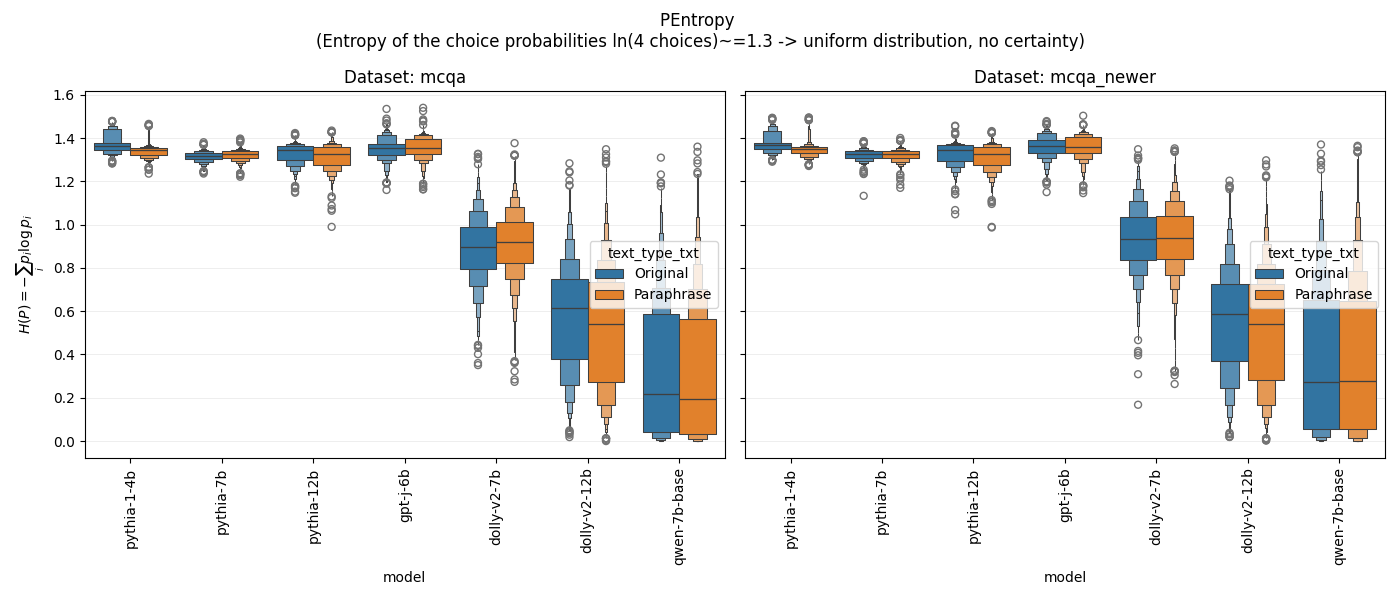}
        \caption{
            \textbf{Entropy of choice probabilities.}
            Box plots showing the distribution of entropy values across different models for original questions and paraphrases. Results are shown for \textit{mcqa} (left) and \textit{mcqa\_newer} (right) datasets. Higher entropy values indicate more uniform probability distributions across answer choices, reflecting greater model uncertainty. Maximum entropy of $ln(4) \approx 1.39$ corresponds to uniform distribution across four choices.
        }
        \label{fig:p_entropy}
\end{figure}

From Figure \ref{fig:p_true}, we analyze the probabilities related to the correct choices, assigned by different models in MCQA and MCQA\_newer. In Figure \ref{fig:p_selected}, we depict the probability assignment to selected choices, comparing the original and paraphrased sentences. Figure \ref{fig:p_entropy}, shows the entropy of the choice probabilities.

The trends appear similar when comparing the two versions of the dataset, with no evidence of verbatim recalls/ memorized patterns that facilitate better causal understanding. Random behavior on "memorized" data - High entropy on MCQA shows no memorization benefit.
A perfect inverse relationship with performance is noted, where pythia/gpt-j/gpt-6b, presents high entropy ($\approx 1.3-1.4$), indicating nearly uniform distributions, which implies no causal understanding. qwen-7b-base, on the other hand presents a low entropy ($\approx0.2-0.6$) indicating confident, decisive predictions. Entropy near $ln(4) \approx 1.39$ for weak models confirms they are essentially guessing randomly. No paraphrase penalty: Original vs Paraphrase performance presents nearly identical behaviours.



\subsection{ Statistical significance tests}
\label{sec:appendix_significance}
We provide statistical tests examining differences in accuracy and entropy between original versus paraphrased sentences, and between MCQA versus MCQA\-newer datasets. T\-tests assess mean differences while ANOVA examines variance across causal relationship types. 
The statistical test results are reported in Tables \ref{tab:significance_onaccuracies} and \ref{tab:significance_onentropies}. These were computed on the accuracies and entropies of models, respectively.

\begin{table*}
  \centering
    \small
    \begin{tabular}{lcccc}
    \toprule
     & \multicolumn{2}{c}{Original vs Paraphrase} & \multicolumn{2}{c}{MCQA vs MCQA-newer}\\
    Model & t-stat & p-value & t-stat & p-value \\
    \midrule
    pythia-1-4b & 0.340 & 0.744 & 0.577 & 0.604 \\
    pythia-7b & -0.184 & 0.859 & 0.412 & 0.708 \\
    pythia-12b & -0.232 & 0.823 & -0.753 & 0.506 \\
    gpt-j-6b & -0.444 & 0.670 & 0.494 & 0.655 \\
    dolly-v2-7b & -0.512 & 0.625 & -0.604 & 0.589 \\
    dolly-v2-12b & -0.669 & 0.525 & -12.427 & 0.001 \\
    qwen-7b-base & 2.032 & 0.082 & -0.468 & 0.672 \\
    \bottomrule
    \end{tabular}
  \caption{\label{tab:significance_onaccuracies}
    T-tests computed on the accuracies for each model between Original and Paraphrase indistinguishable of the dateset and, between the original sentences of MCQA and the original sentences of MCQA-newer. Accuracies feed to the t-tests were the means of the binary correctness grouped by the causal relationship type.
  }
\end{table*}

\begin{table*}
  \centering
    \small
    \begin{tabular}{lrrrrrrrr}
    \toprule
     & \multicolumn{3}{c}{Original vs Paraphrase} & \multicolumn{3}{c}{MCQA vs MCQA-newer} & \multicolumn{2}{c}{Causal-type ANOVA}\\
    Model & t-stat & p-value & effect size & t-stat & p-value & effect size & f-stat & p-value \\
    \midrule

pythia-1-4b & 69.273 & 0.000 & 0.884 & 0.563 & 0.573 & 0.015 & 15.509 & 0.000 \\
pythia-7b & -10.205 & 0.000 & -0.152 & -11.637 & 0.000 & -0.307 & 25.805 & 0.000 \\
pythia-12b & 18.441 & 0.000 & 0.286 & 0.655 & 0.513 & 0.017 & 8.400 & 0.000 \\
gpt-j-6b & 0.270 & 0.787 & 0.004 & -7.847 & 0.000 & -0.207 & 139.133 & 0.000 \\
dolly-v2-7b & -5.656 & 0.000 & -0.083 & -11.328 & 0.000 & -0.301 & 40.102 & 0.000 \\
dolly-v2-12b & 11.673 & 0.000 & 0.181 & 2.846 & 0.004 & 0.076 & 25.551 & 0.000 \\
qwen-7b-base & 1.541 & 0.123 & 0.023 & -5.679 & 0.000 & -0.150 & 153.846 & 0.000 \\

    \bottomrule
    \end{tabular}
  \caption{\label{tab:significance_onentropies}
    Statistical tests computed on the entropies for each model between Original and Paraphrase indistinguishable of the dateset and, between the original sentences of MCQA and the original sentences of MCQA-newer.
  }
\end{table*}




\subsection{Consistency analysis} \label{app:consistency}

\begin{figure}[hbt]
    \centering
    \includegraphics[width=1.0\linewidth]{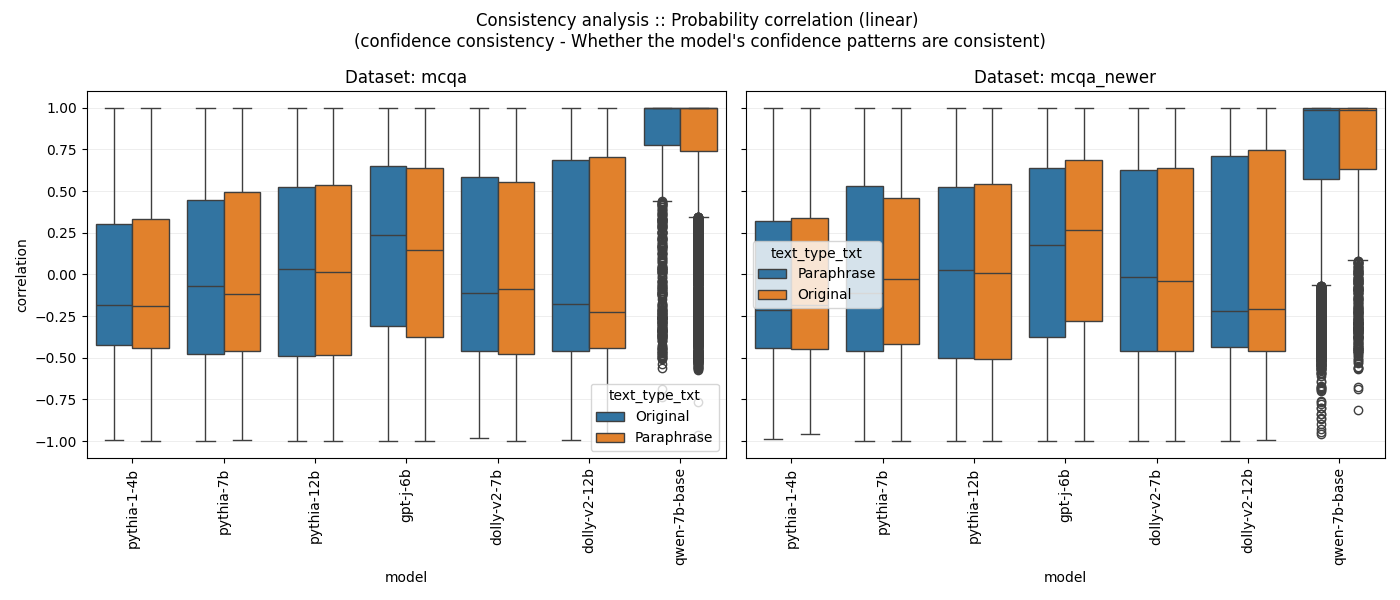}
    \caption{Probability correlation: confidence consistency — whether the model's confidence patterns are consistent.}
    \label{fig:consistency_corr}
\end{figure}

\begin{figure}[hbt]
    \centering
        \centering
        \includegraphics[width=1.0\linewidth]{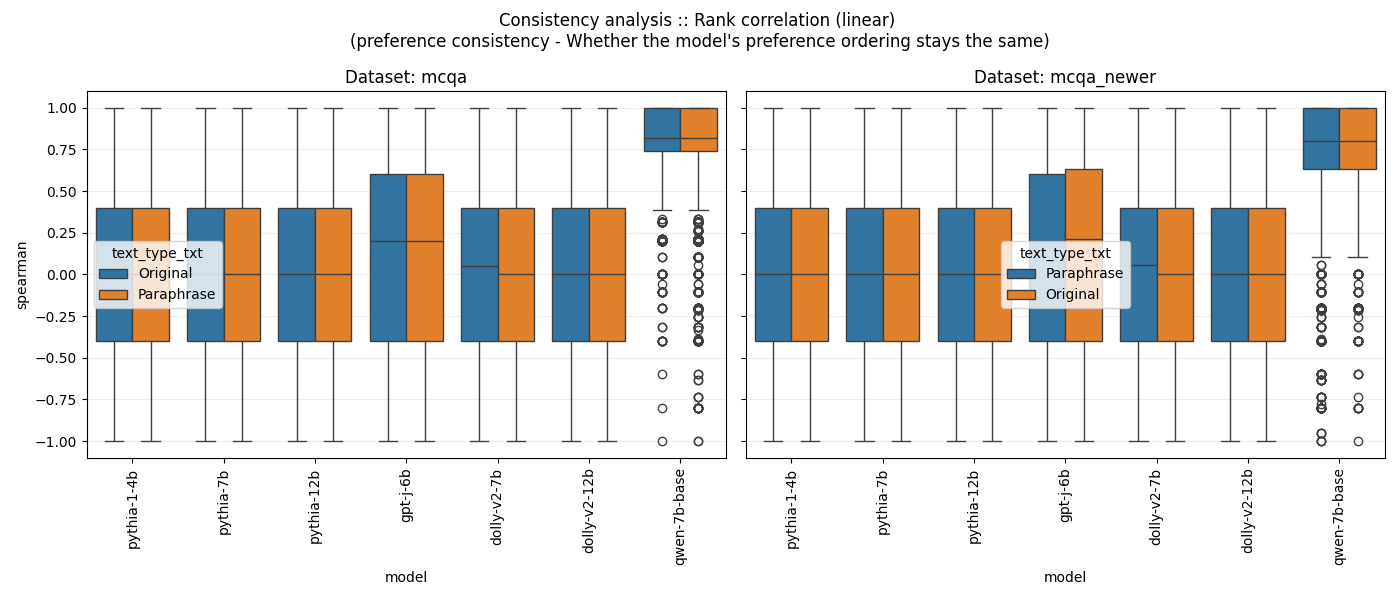}
        \caption{Rank correlation: preference consistency - Whether the model's preference ordering stays the same.}
        \label{fig:consistency_spear}
\end{figure}

\begin{figure}[hbt]
    \centering
        \centering
        \includegraphics[width=1.0\linewidth]{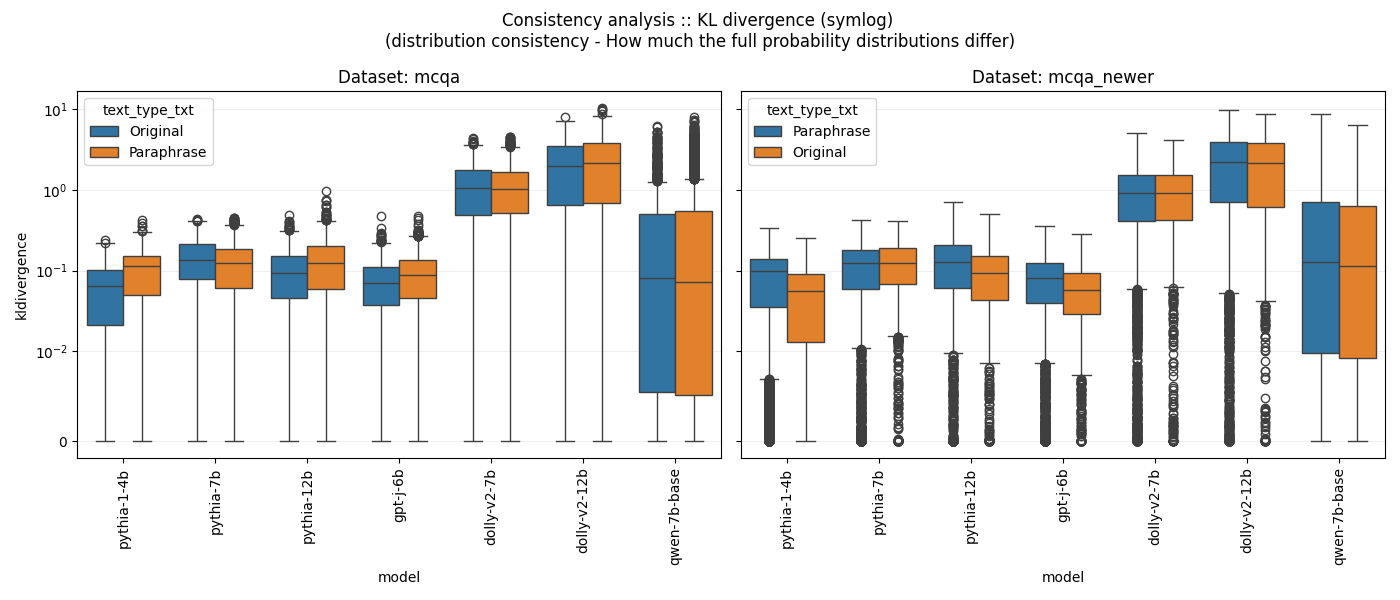}
        \caption{KL divergence: distribution consistency - How much the full probability distributions differ.}
        \label{fig:consistency_kldiv}
\end{figure}

For each sentence with multiple probes, we computed Pearson correlation between choice probability vectors to assess confidence consistency, Spearman rank correlation to evaluate preference consistency by comparing ranked orders, and KL divergence to quantify overall distributional shifts between probabilistic outputs. Tables \ref{tab:agg_consistency}
and Figures \ref{fig:consistency_spear}, \ref{fig:consistency_corr} and \ref{fig:consistency_kldiv}.
High consistency would indicate robust causal understanding, while low consistency suggests models treat paraphrases as unrelated inputs.


---------- 




\begin{figure*}[hbt]
    \centering
    \begin{minipage}{0.96\textwidth}
        \centering
        \includegraphics[width=\linewidth]{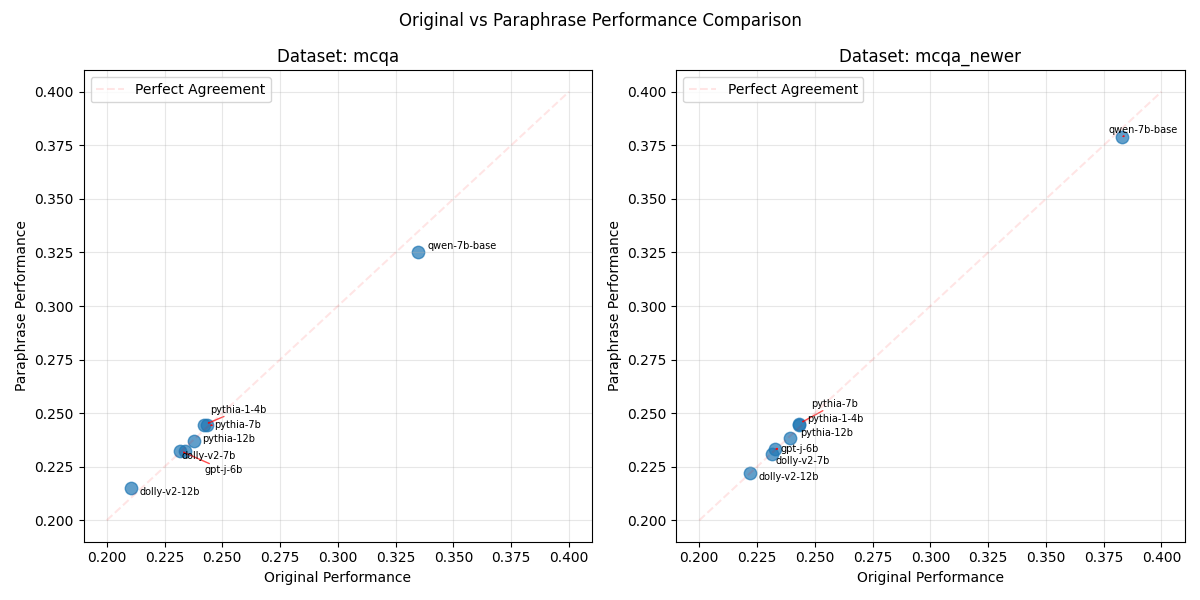}
        \caption{Accuracy comparison of original vs paraphrase in \emph{mcqa} (left) and \emph{mcqa\_newer} (right).}
        \label{fig:performance_originalVSparaphrase}
    \end{minipage}
\end{figure*}

\begin{table*}
  \centering
    \begin{tabular}{lllrrr}
    \toprule
    model & dataset & text type & prob correlation & Spearman & KL div \\
    \midrule
    pythia-1-4b & mcqa & Paraphrase & -0.007 & 0.003 & 0.105 \\
    pythia-1-4b & mcqa\_newer & Paraphrase & -0.020 & -0.009 & 0.093 \\
    pythia-1-4b & mcqa\_newer & Original & -0.009 & 0.003 & 0.058 \\
    pythia-1-4b & mcqa & Original & 0.002 & 0.025 & 0.066 \\
    \hline
    pythia-7b & mcqa\_newer & Paraphrase & 0.003 & -0.001 & 0.127 \\
    pythia-7b & mcqa\_newer & Original & 0.002 & 0.000 & 0.130 \\
    pythia-7b & mcqa & Paraphrase & -0.009 & -0.014 & 0.129 \\
    pythia-7b & mcqa & Original & -0.017 & -0.026 & 0.147 \\
    \hline
    pythia-12b & mcqa & Paraphrase & 0.019 & -0.001 & 0.137 \\
    pythia-12b & mcqa & Original & 0.025 & 0.011 & 0.105 \\
    pythia-12b & mcqa\_newer & Paraphrase & 0.016 & -0.005 & 0.143 \\
    pythia-12b & mcqa\_newer & Original & 0.024 & 0.005 & 0.108 \\
    \hline
    gpt-j-6b & mcqa\_newer & Paraphrase & 0.123 & 0.113 & 0.087 \\
    
    gpt-j-6b & mcqa & Original & 0.156 & 0.119 & 0.078 \\
    gpt-j-6b & mcqa\_newer & Original & 0.183 & 0.160 & 0.066 \\
    gpt-j-6b & mcqa & Paraphrase & 0.109 & 0.098 & 0.093 \\
    \hline
    dolly-v2-7b & mcqa\_newer & Paraphrase & 0.059 & 0.040 & 1.049 \\
    dolly-v2-7b & mcqa\_newer & Original & 0.054 & 0.045 & 1.045 \\
    dolly-v2-7b & mcqa & Paraphrase & 0.018 & 0.017 & 1.141 \\
    dolly-v2-7b & mcqa & Original & 0.029 & 0.026 & 1.186 \\
    \hline
    dolly-v2-12b & mcqa & Original & 0.050 & 0.006 & 2.240 \\
    dolly-v2-12b & mcqa\_newer & Paraphrase & 0.053 & 0.026 & 2.475 \\
    dolly-v2-12b & mcqa\_newer & Original & 0.056 & 0.026 & 2.430 \\
    dolly-v2-12b & mcqa & Paraphrase & 0.045 & 0.015 & 2.376 \\
    \hline
    qwen-7b-base & mcqa & Original & 0.777 & 0.775 & 0.494 \\
    qwen-7b-base & mcqa & Paraphrase & 0.762 & 0.769 & 0.527 \\
    qwen-7b-base & mcqa\_newer & Original & 0.727 & 0.739 & 0.547 \\
    qwen-7b-base & mcqa\_newer & Paraphrase & 0.710 & 0.727 & 0.602 \\
    \bottomrule
    \end{tabular}    
  \caption{\label{tab:agg_consistency}
    Consistency analysis over predictions aggregated by model, dataset and text type.     
  }
\end{table*}

\subsection{Overconfidence Analysis} \label{app:overconfidence}
\begin{figure}[hbt]
    \centering
        \centering
        \includegraphics[width=1.0\linewidth]{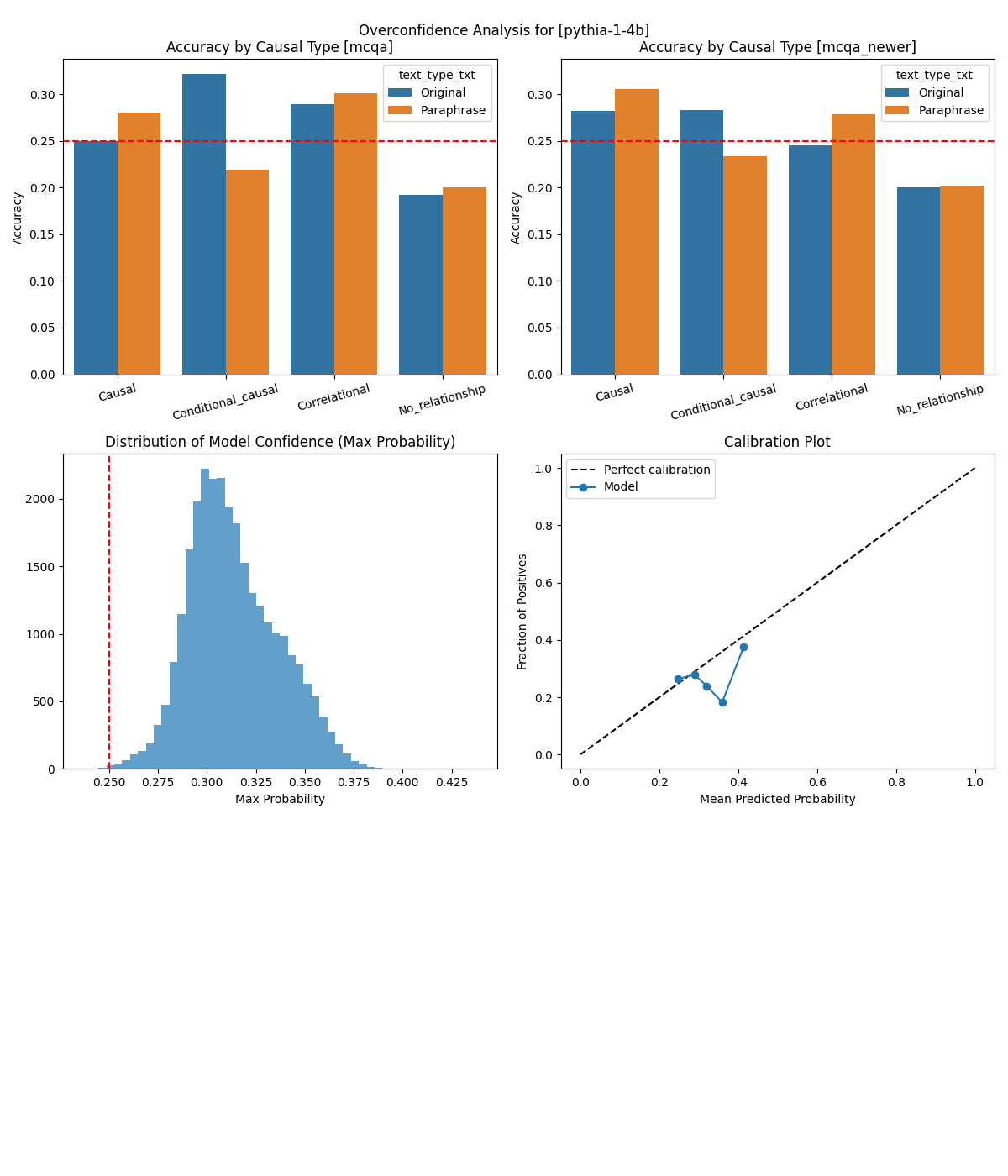}
        \caption{
            \textbf{Overconfidence analysis for Pythia 1.4B.}
            \emph{Top row:} Accuracy by causal relationship type for original questions and paraphrases in \textit{mcqa} (left) and \textit{mcqa\_newer} (right).
            \emph{Middle row:} Model confidence distribution using 50 bins (left) and calibration plot showing predicted probabilities (x-axis) versus actual accuracy (y-axis) with data grouped into 20 bins (right).
        }        
        \label{fig:overconfidence_pythia-1-4b}
\end{figure}

\begin{figure}[hbt]
    \centering
        \centering
        \includegraphics[width=1.0\linewidth]{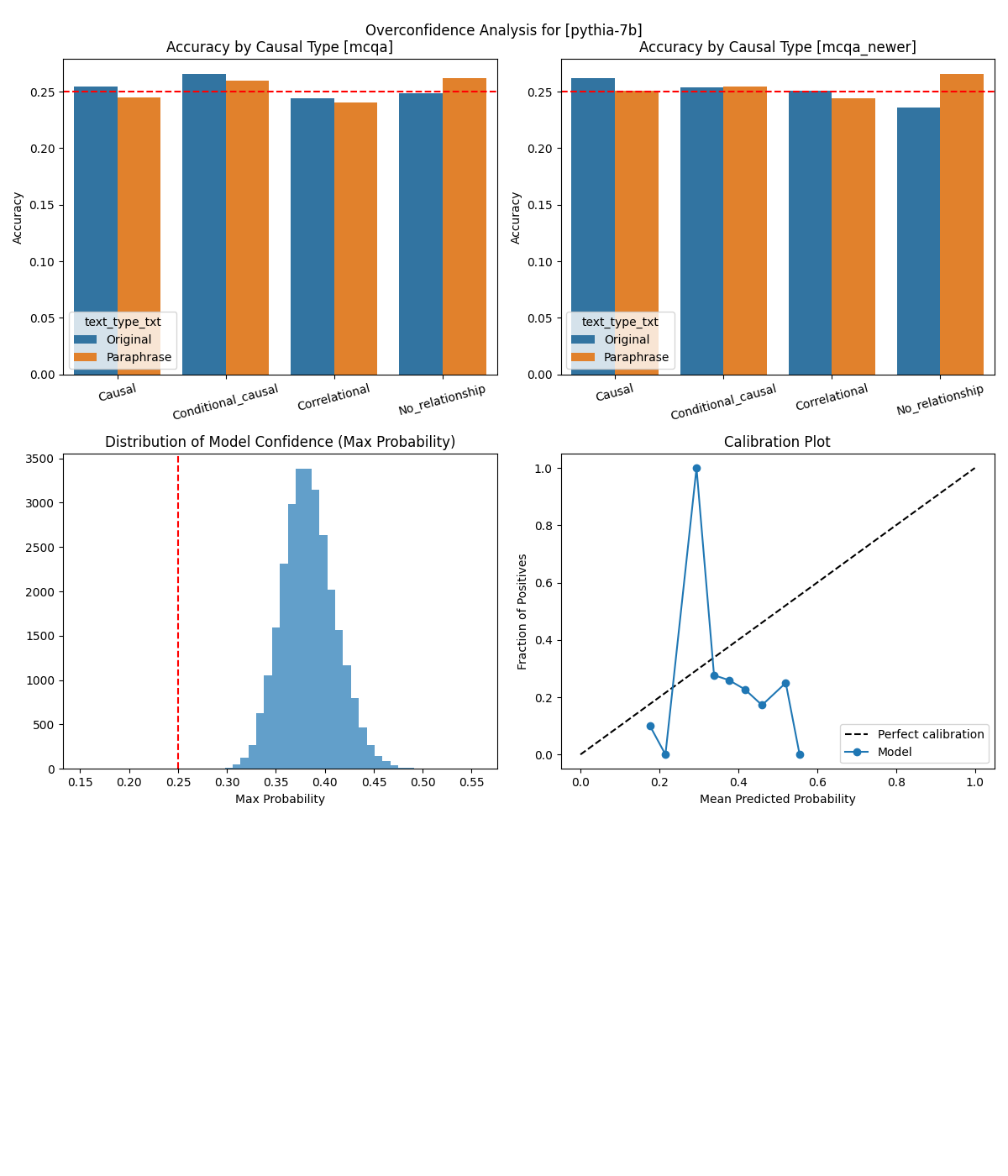}
        \caption{
            \textbf{Overconfidence analysis for Pythia 7B.}
            \emph{Top row:} Accuracy by causal relationship type for original questions and paraphrases in \textit{mcqa} (left) and \textit{mcqa\_newer} (right).
            \emph{Middle row:} Model confidence distribution using 50 bins (left) and calibration plot showing predicted probabilities (x-axis) versus actual accuracy (y-axis) with data grouped into 20 bins (right).
        }                
        \label{fig:overconfidence_pythia-7b}
\end{figure}

\begin{figure}[hbt]
    \centering
        \centering
        \includegraphics[width=1.0\linewidth]{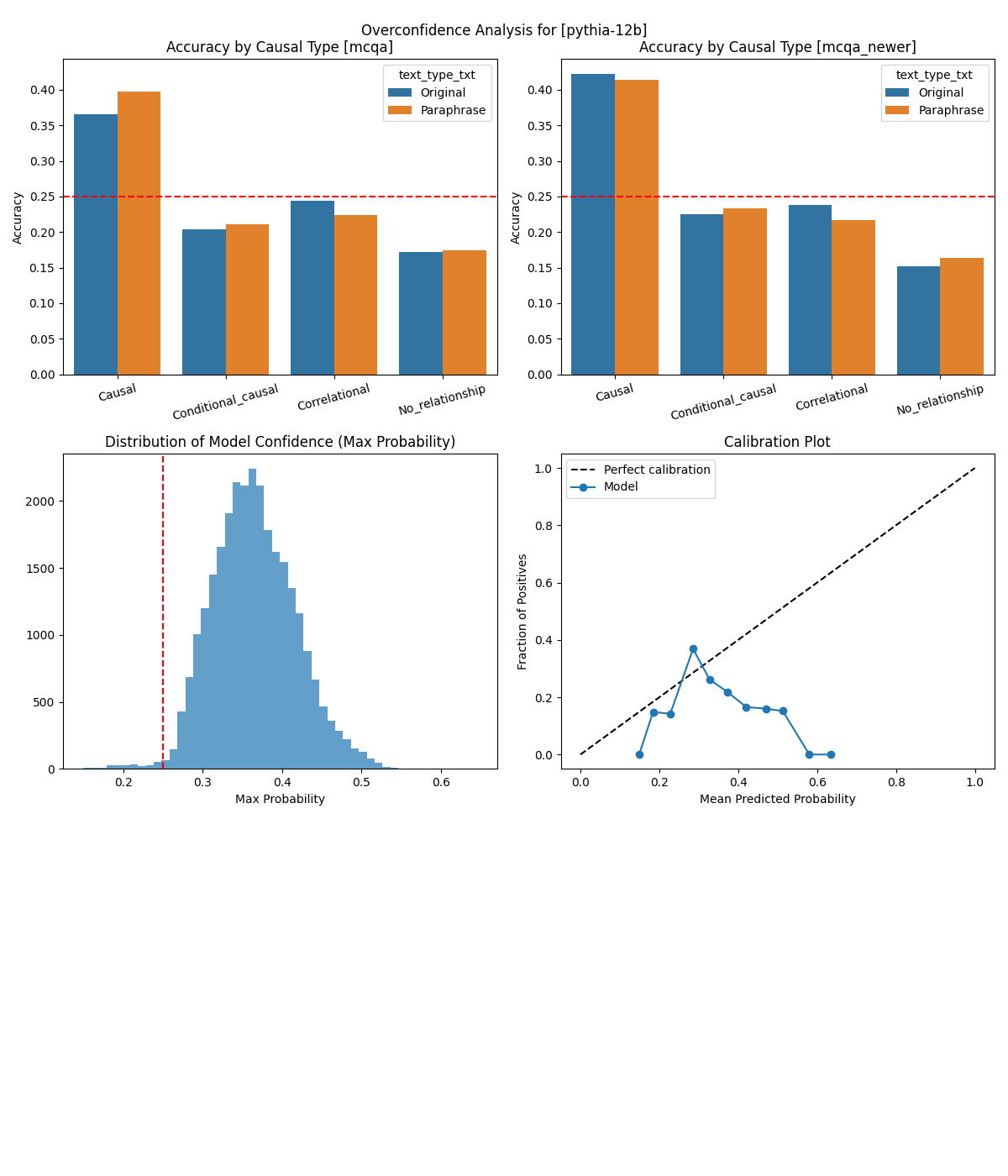}        
        \caption{
            \textbf{Overconfidence analysis for Pythia 12B.}
            \emph{Top row:} Accuracy by causal relationship type for original questions and paraphrases in \textit{mcqa} (left) and \textit{mcqa\_newer} (right).
            \emph{Middle row:} Model confidence distribution using 50 bins (left) and calibration plot showing predicted probabilities (x-axis) versus actual accuracy (y-axis) with data grouped into 20 bins (right).
        }                
        \label{fig:overconfidence_pythia-12b}
\end{figure}

\begin{figure}[hbt]
    \centering
        \centering
        \includegraphics[width=1.0\linewidth]{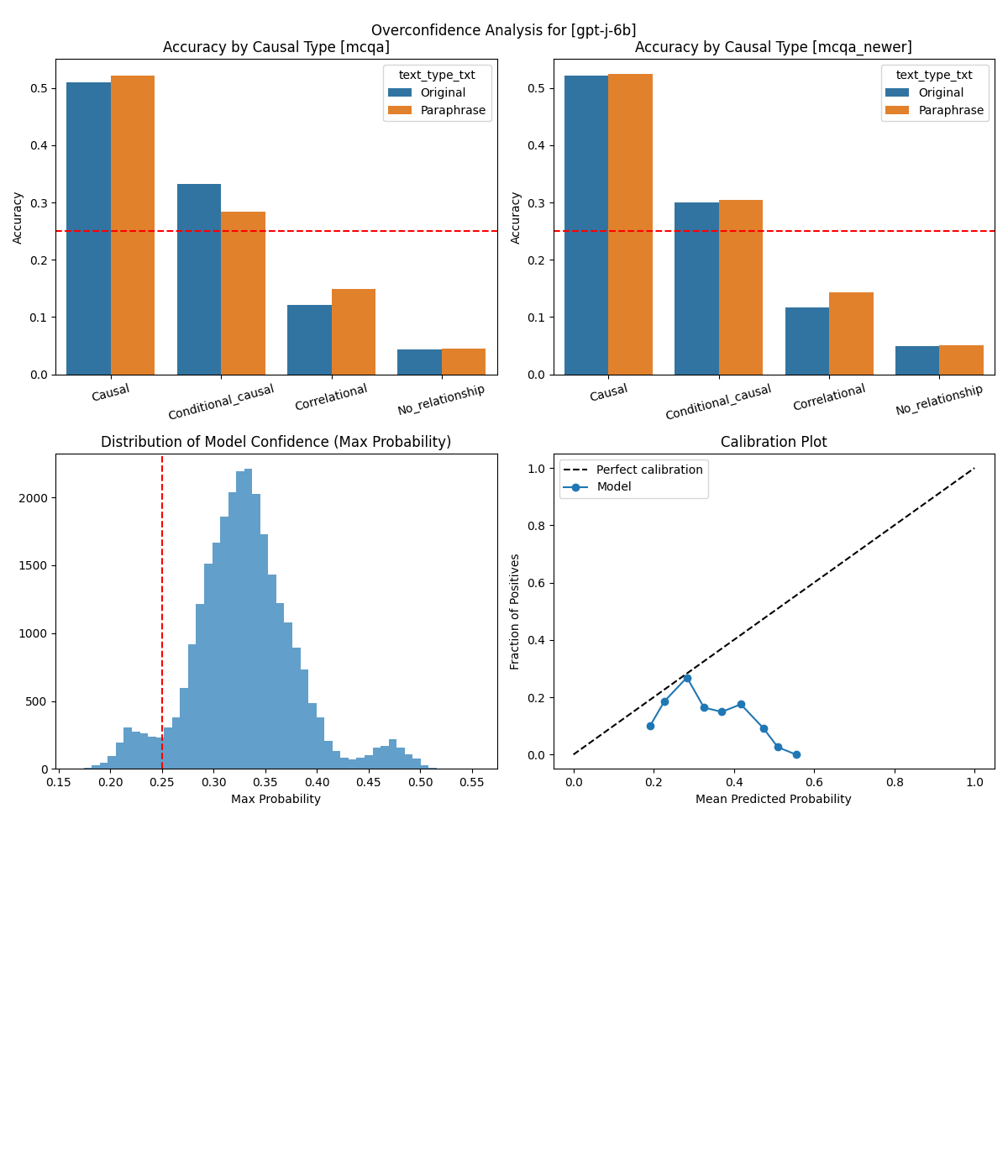}
        \caption{
            \textbf{Overconfidence analysis for GPT-j 6B.}
            \emph{Top row:} Accuracy by causal relationship type for original questions and paraphrases in \textit{mcqa} (left) and \textit{mcqa\_newer} (right).
            \emph{Middle row:} Model confidence distribution using 50 bins (left) and calibration plot showing predicted probabilities (x-axis) versus actual accuracy (y-axis) with data grouped into 20 bins (right).
        }        
        \label{fig:overconfidence_gpt-j-6b}
\end{figure}

\begin{figure}[hbt]
    \centering
        \centering
        \includegraphics[width=1.0\linewidth]{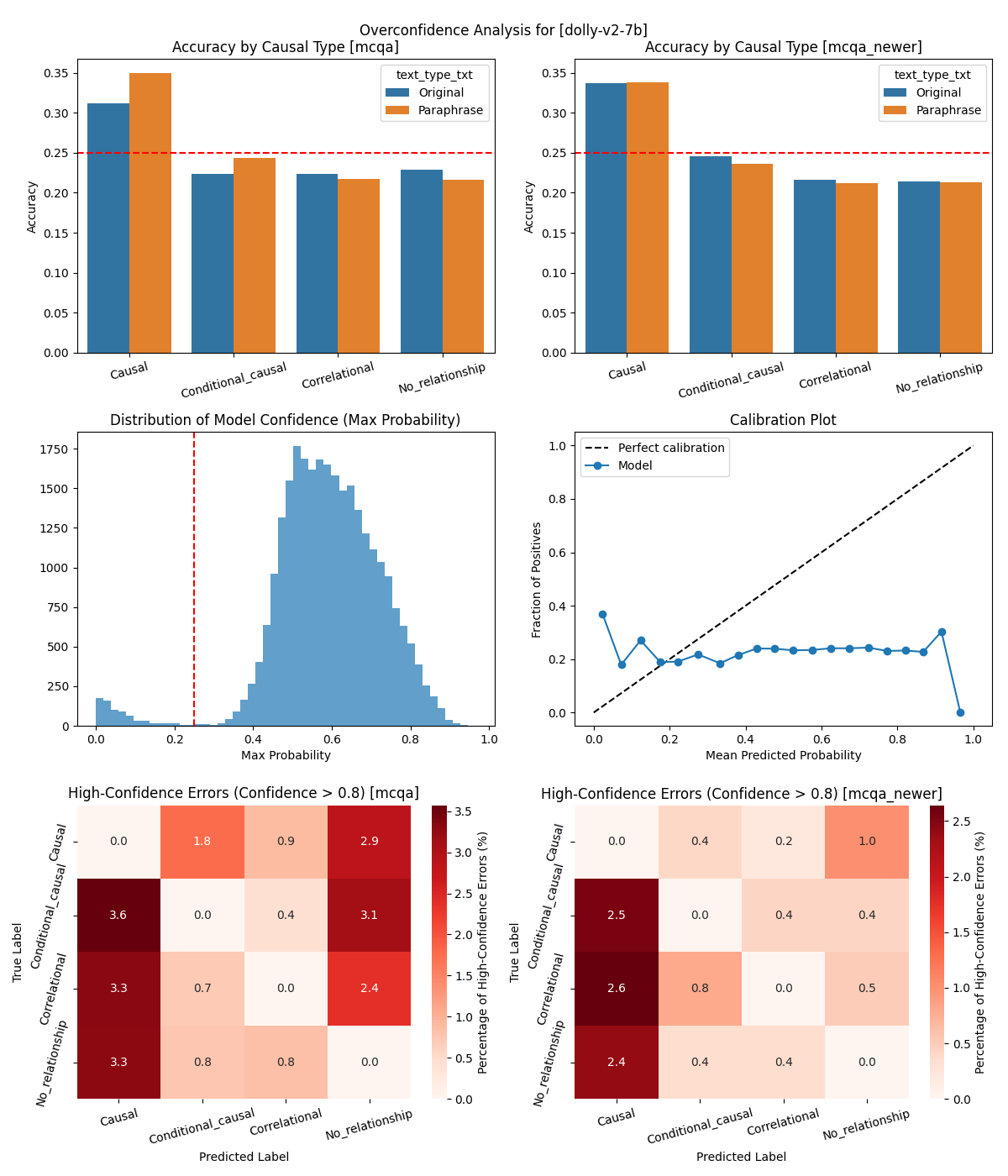}
        \caption{
            \textbf{Overconfidence analysis for Dolly-v12 7B.}
            \emph{Top row:} Accuracy by causal relationship type for original questions and paraphrases in \textit{mcqa} (left) and \textit{mcqa\_newer} (right).
            \emph{Middle row:} Model confidence distribution using 50 bins (left) and calibration plot showing predicted probabilities (x-axis) versus actual accuracy (y-axis) with data grouped into 20 bins (right).
            \emph{Bottom row:} Confusion matrices for high-confidence errors (confidence >0.8) in \textit{mcqa} (left) and \textit{mcqa\_newer} (right). Heatmap values represent the percentage of each true class that was misclassified with high confidence.
        }
        \label{fig:overconfidence_dolly-7b}
\end{figure}

\begin{figure}[hbt]
    \centering
        \centering
        \includegraphics[width=1.0\linewidth]{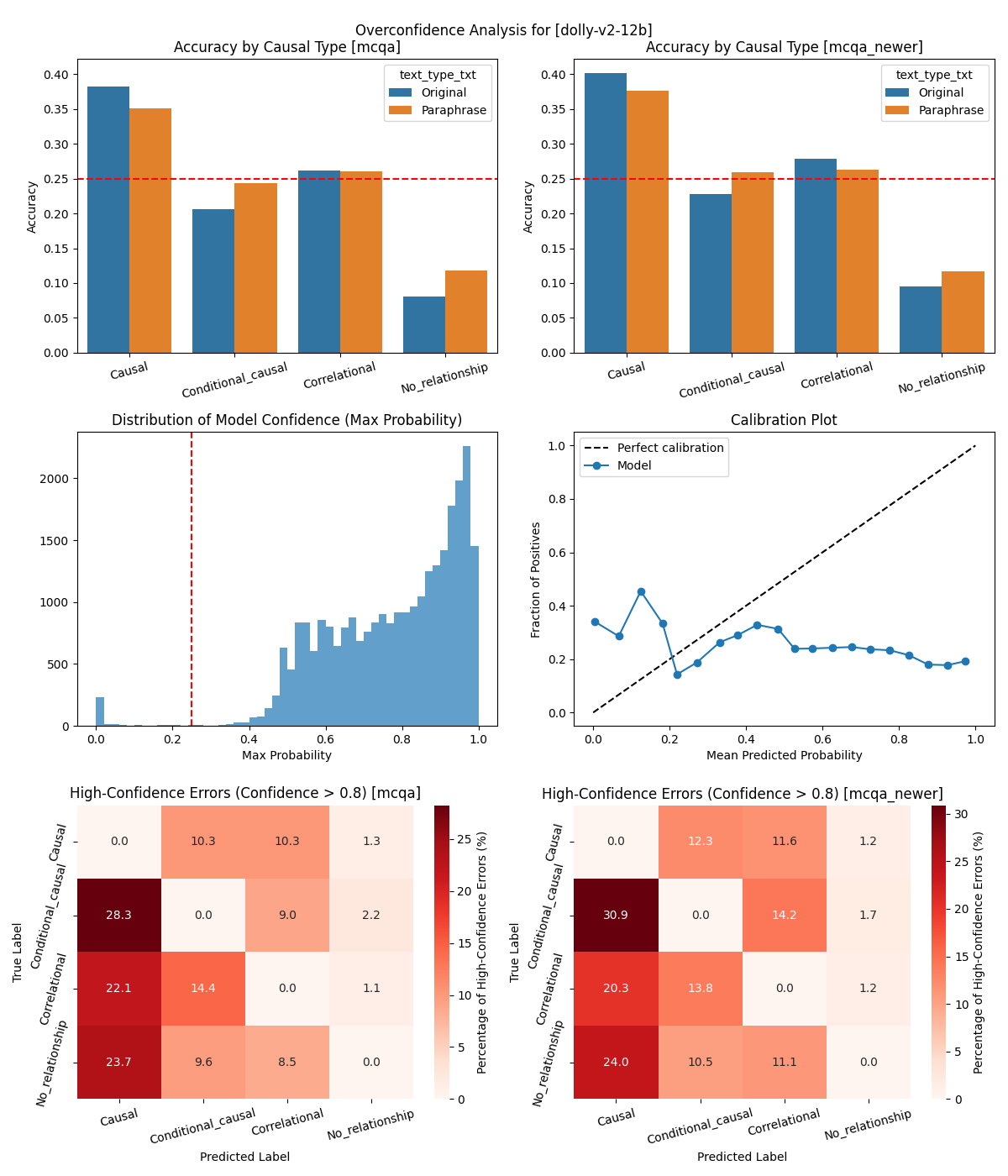}
        \caption{
            \textbf{Overconfidence analysis for olly-v12 12B.}
            \emph{Top row:} Accuracy by causal relationship type for original questions and paraphrases in \textit{mcqa} (left) and \textit{mcqa\_newer} (right).
            \emph{Middle row:} Model confidence distribution using 50 bins (left) and calibration plot showing predicted probabilities (x-axis) versus actual accuracy (y-axis) with data grouped into 20 bins (right).
            \emph{Bottom row:} Confusion matrices for high-confidence errors (confidence >0.8) in \textit{mcqa} (left) and \textit{mcqa\_newer} (right). Heatmap values represent the percentage of each true class that was misclassified with high confidence.
        }        
        \label{fig:overconfidence_dolly-12b}
\end{figure}

\begin{figure}[hbt]
    \centering
        \centering
        \includegraphics[width=1.0\linewidth]{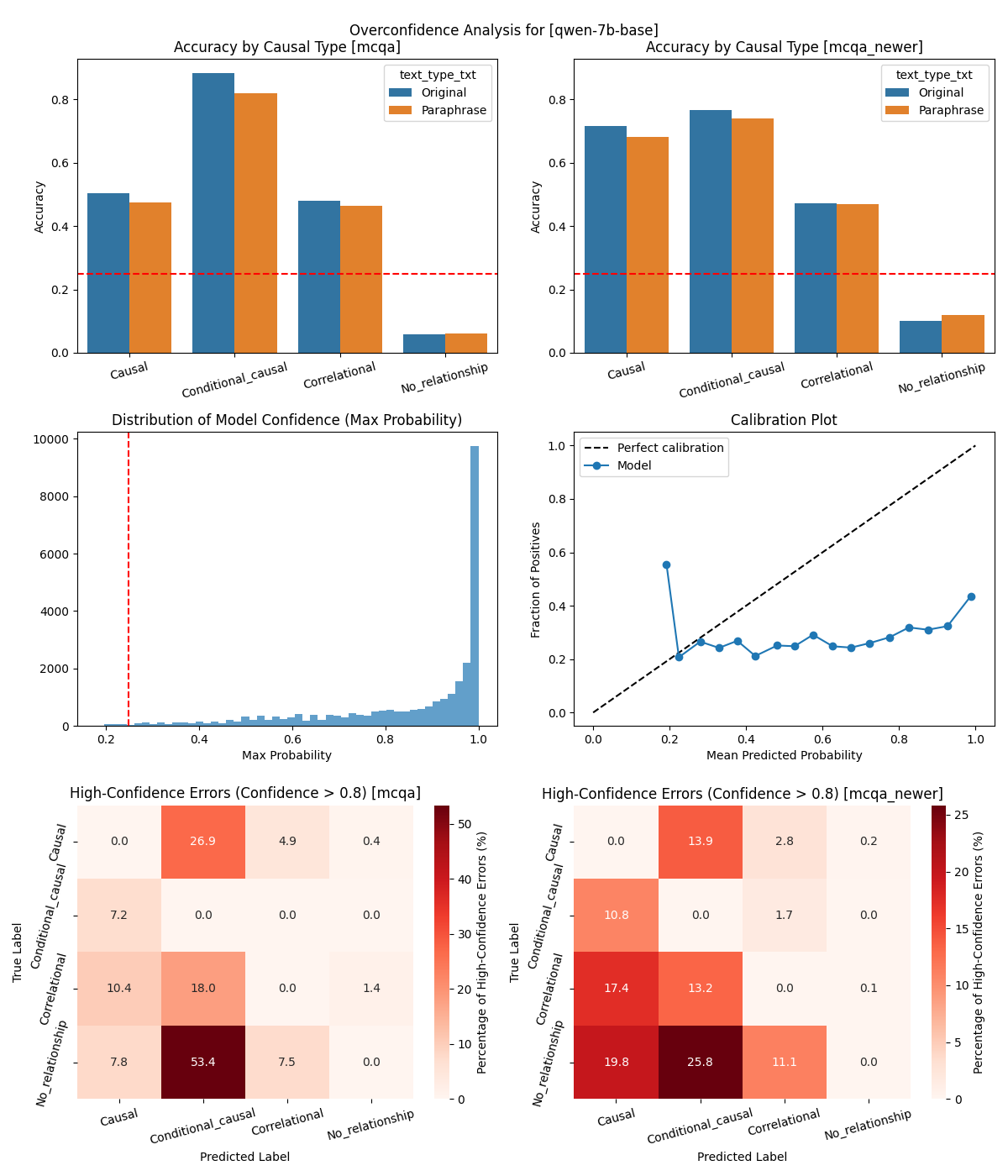}
        \caption{
            \textbf{Overconfidence analysis for Qwen 7B.}
            \emph{Top row:} Accuracy by causal relationship type for original questions and paraphrases in \textit{mcqa} (left) and \textit{mcqa\_newer} (right).
            \emph{Middle row:} Model confidence distribution using 50 bins (left) and calibration plot showing predicted probabilities (x-axis) versus actual accuracy (y-axis) with data grouped into 20 bins (right).
            \emph{Bottom row:} Confusion matrices for high-confidence errors (confidence >0.8) in \textit{mcqa} (left) and \textit{mcqa\_newer} (right). Heatmap values represent the percentage of each true class that was misclassified with high confidence.
        }                
        \label{fig:overconfidence_qwen-7b}
\end{figure}

We present calibration analyses (Figures \ref{fig:overconfidence_pythia-1-4b}-\ref{fig:overconfidence_qwen-7b}) for each model, examining the relationship between predicted confidence and actual accuracy. Each model's analysis includes accuracy breakdowns by causal type, confidence distributions, calibration plots, and confusion matrices for high-confidence errors. 
These reveal overconfidence patterns, particularly in instruction-tuned models.

\subsection{Uncertainty Analysis}\label{app:uncertainty}
\begin{figure}[hbt]
    \centering
        \centering
        \includegraphics[width=1.0\linewidth]{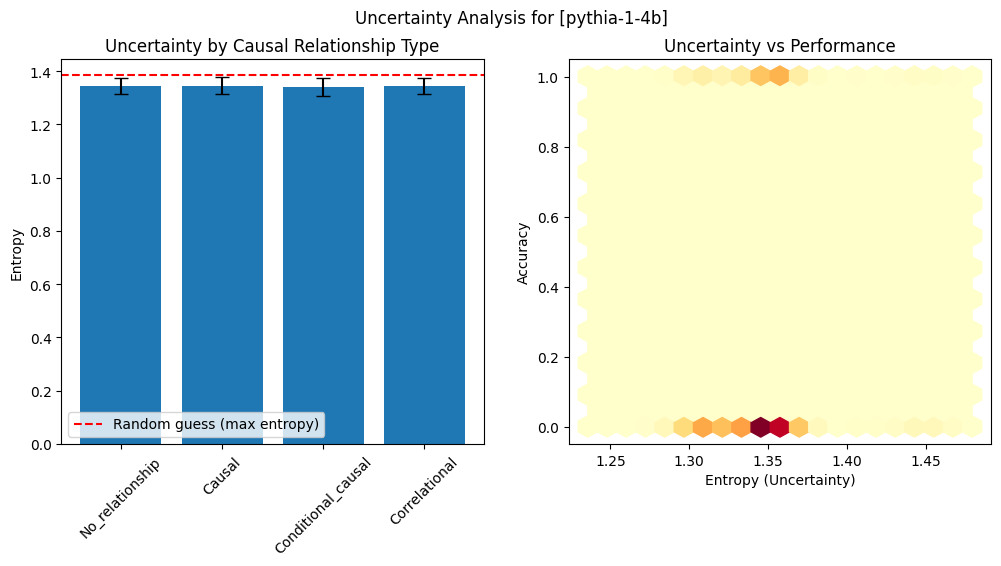}
        \caption{
            \textbf{Uncertainty analysis for Pythia 1.4B.}
            \emph{Left:} Entropy-based uncertainty by causal relationship type. The red dashed line indicates maximum entropy (random guessing baseline).
            \emph{Right:}  Relationship between model uncertainty (entropy) and prediction accuracy, showing how confidence relates to performance.
        }        
        \label{fig:uncertainty_pythia-1-4b}
\end{figure}

\begin{figure}[hbt]
    \centering
        \centering
        \includegraphics[width=1.0\linewidth]{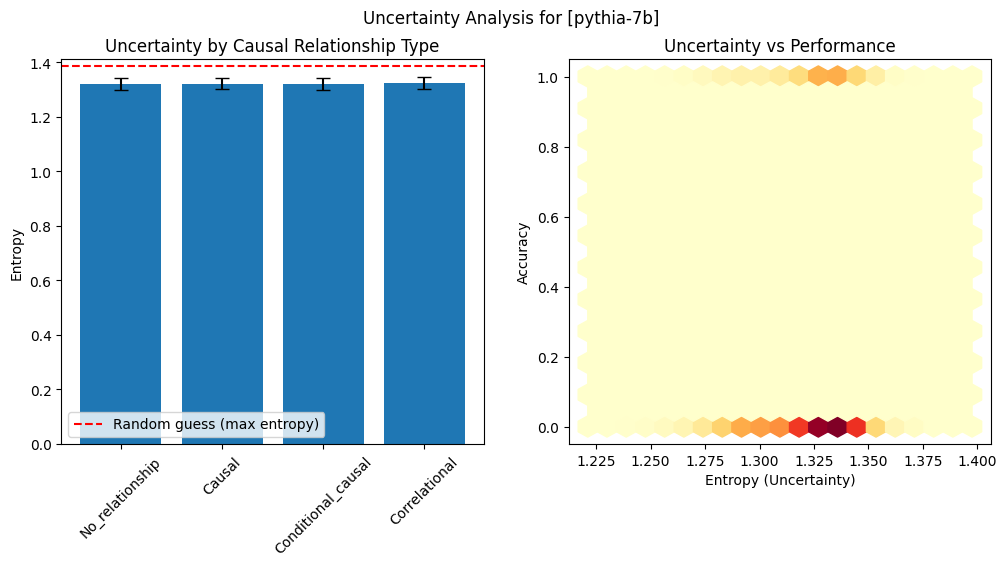}
        \caption{
            \textbf{Uncertainty analysis for Pythia 7B.}
            \emph{Left:} Entropy-based uncertainty by causal relationship type. The red dashed line indicates maximum entropy (random guessing baseline).
            \emph{Right:}  Relationship between model uncertainty (entropy) and prediction accuracy, showing how confidence relates to performance.
        }                
        \label{fig:uncertainty_pythia-7b}
\end{figure}

\begin{figure}[hbt]
    \centering
        \centering
        \includegraphics[width=1.0\linewidth]{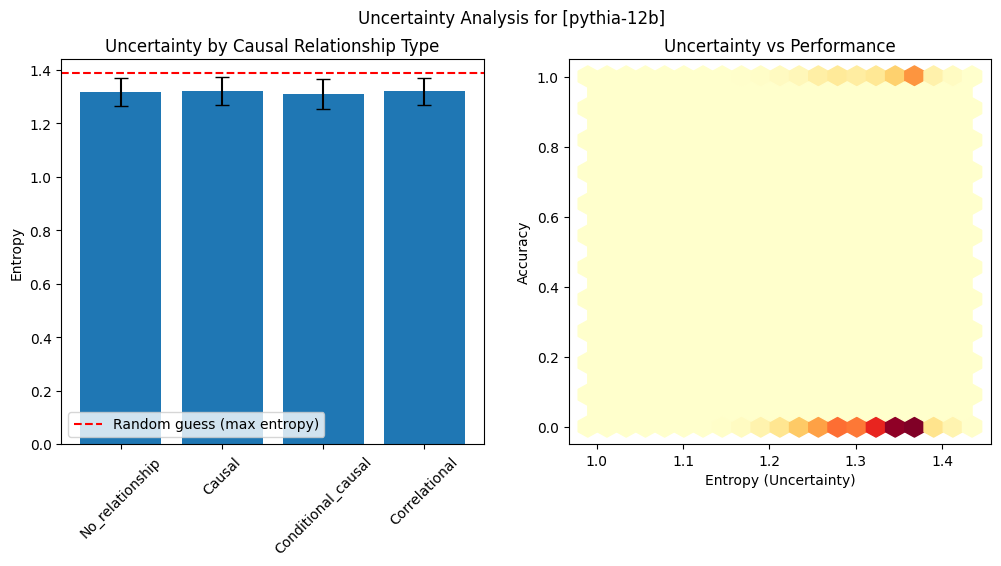}        
        \caption{
            \textbf{Uncertainty analysis for Pythia 12B.}
            \emph{Left:} Entropy-based uncertainty by causal relationship type. The red dashed line indicates maximum entropy (random guessing baseline).
            \emph{Right:}  Relationship between model uncertainty (entropy) and prediction accuracy, showing how confidence relates to performance.
        }                
        \label{fig:uncertainty_pythia-12b}
\end{figure}

\begin{figure}[hbt]
    \centering
        \centering
        \includegraphics[width=1.0\linewidth]{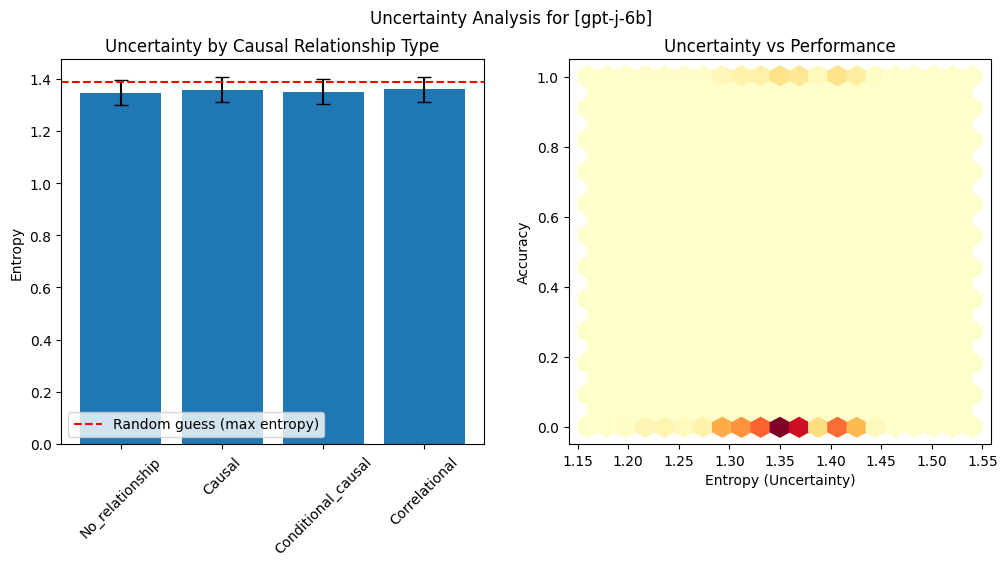}
        \caption{
            \textbf{Uncertainty analysis for GPT-j 6B.}
            \emph{Left:} Entropy-based uncertainty by causal relationship type. The red dashed line indicates maximum entropy (random guessing baseline).
            \emph{Right:}  Relationship between model uncertainty (entropy) and prediction accuracy, showing how confidence relates to performance.
        }        
        \label{fig:uncertainty_gpt-j-6b}
\end{figure}

\begin{figure}[hbt]
    \centering
        \centering
        \includegraphics[width=1.0\linewidth]{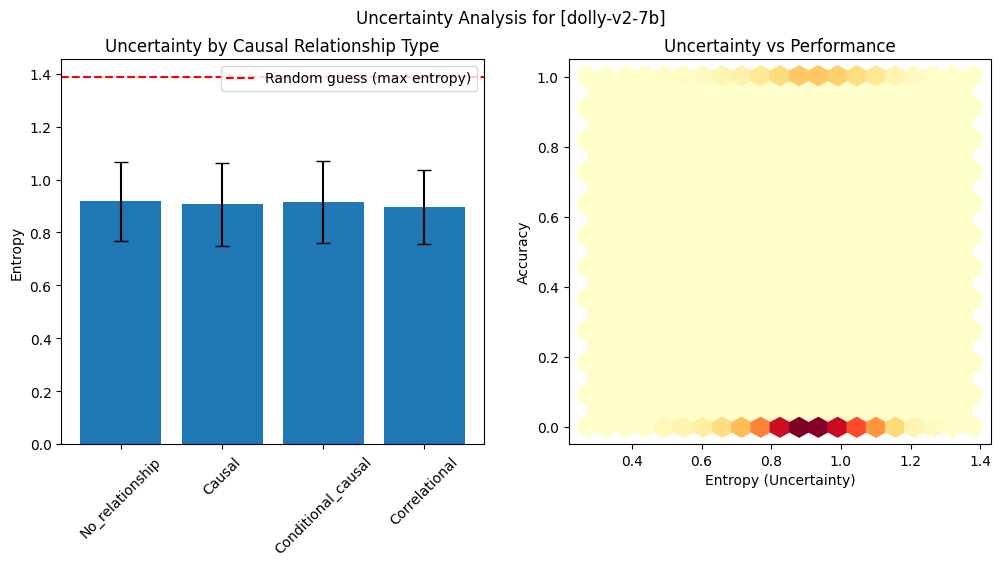}
        \caption{
            \textbf{Uncertainty analysis for Dolly-v12 7B.}
            \emph{Left:} Entropy-based uncertainty by causal relationship type. The red dashed line indicates maximum entropy (random guessing baseline).
            \emph{Right:}  Relationship between model uncertainty (entropy) and prediction accuracy, showing how confidence relates to performance.
        }
        \label{fig:uncertainty_dolly-7b}
\end{figure}

\begin{figure}[hbt]
    \centering
        \centering
        \includegraphics[width=1.0\linewidth]{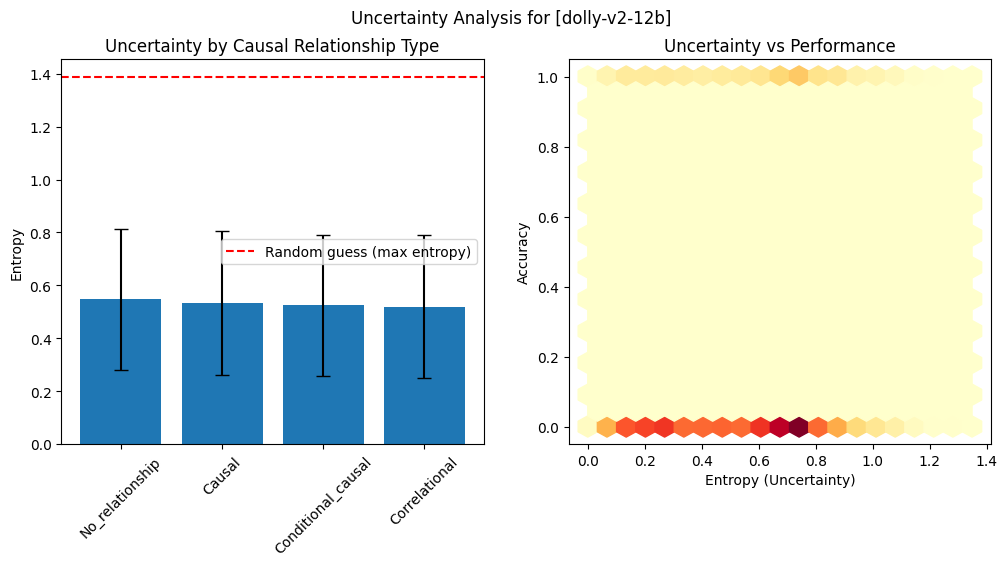}
        \caption{
            \textbf{uncertainty analysis for olly-v12 12B.}
            \emph{Left:} Entropy-based uncertainty by causal relationship type. The red dashed line indicates maximum entropy (random guessing baseline).
            \emph{Right:}  Relationship between model uncertainty (entropy) and prediction accuracy, showing how confidence relates to performance.
        }        
        \label{fig:uncertainty_dolly-12b}
\end{figure}

\begin{figure}[hbt]
    \centering
        \centering
        \includegraphics[width=1.0\linewidth]{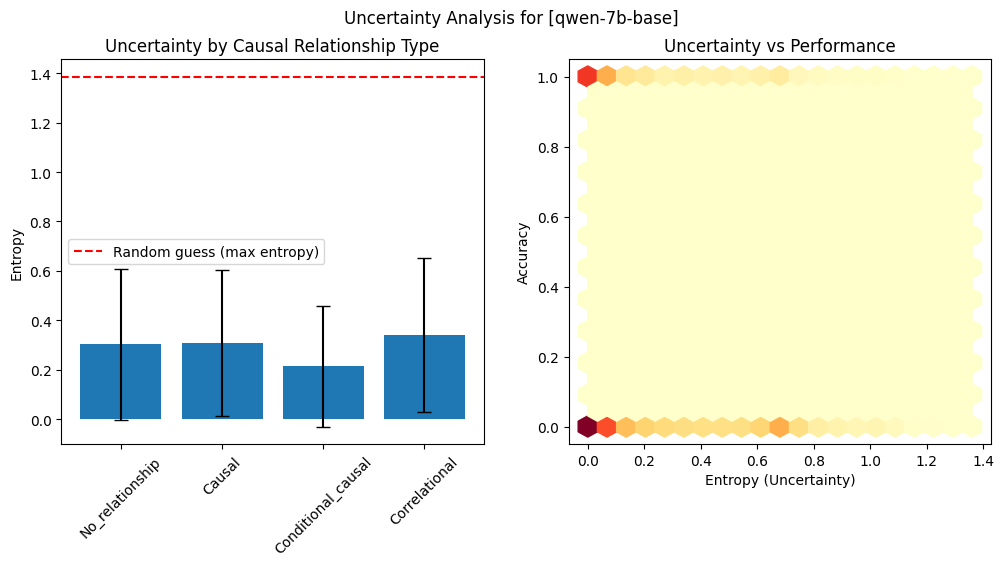}
        \caption{
            \textbf{Uncertainty analysis for Qwen 7B.}
            \emph{Left:} Entropy-based uncertainty by causal relationship type. The red dashed line indicates maximum entropy (random guessing baseline).
            \emph{Right:}  Relationship between model uncertainty (entropy) and prediction accuracy, showing how confidence relates to performance.
        }                
        \label{fig:uncertainty_qwen-7b}
\end{figure}

This section examines (Figures \ref{fig:uncertainty_pythia-1-4b} - \ref{fig:uncertainty_qwen-7b}) how uncertainty (measured by entropy) varies across different causal relationship types and its correlation with model accuracy. The analysis reveals that conditional causal relationships consistently induce the highest uncertainty across all models, suggesting limitations in compositional causal reasoning rather than simple memorization effects.



\section{Task 2: Verbatim Memorization Probing} \label{app:task2}

\begin{figure}[hbt]
    \centering
        \centering
        \includegraphics[width=1.0\linewidth]{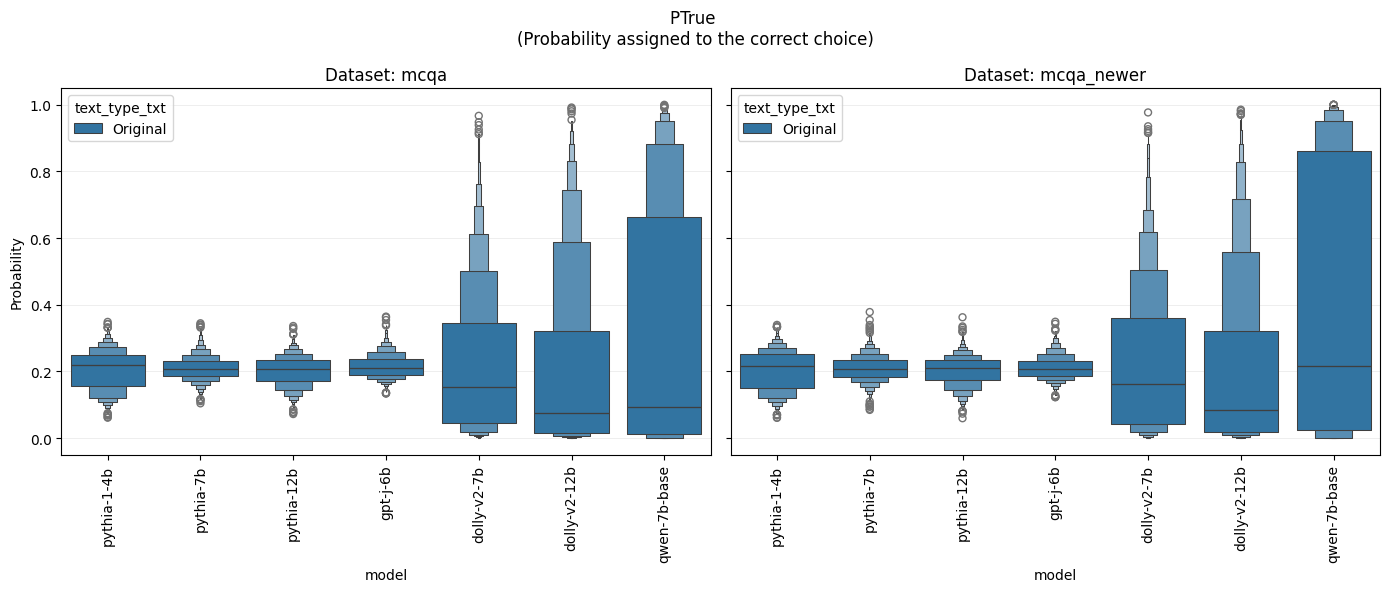}
        \caption{
            \textbf{Probabilities assigned to the correct choice.}
            Box plots showing the distribution of probabilities assigned to correct answers by different models for original questions and paraphrases. Results are shown for \textit{mcqa} (left) and \textit{mcqa\_newer} (right) datasets. Higher probabilities indicate greater model confidence in correct predictions.            
        }
        \label{fig:task2_p_true}
\end{figure}

\begin{figure}[hbt]
    \centering
        \centering
        \includegraphics[width=1.0\linewidth]{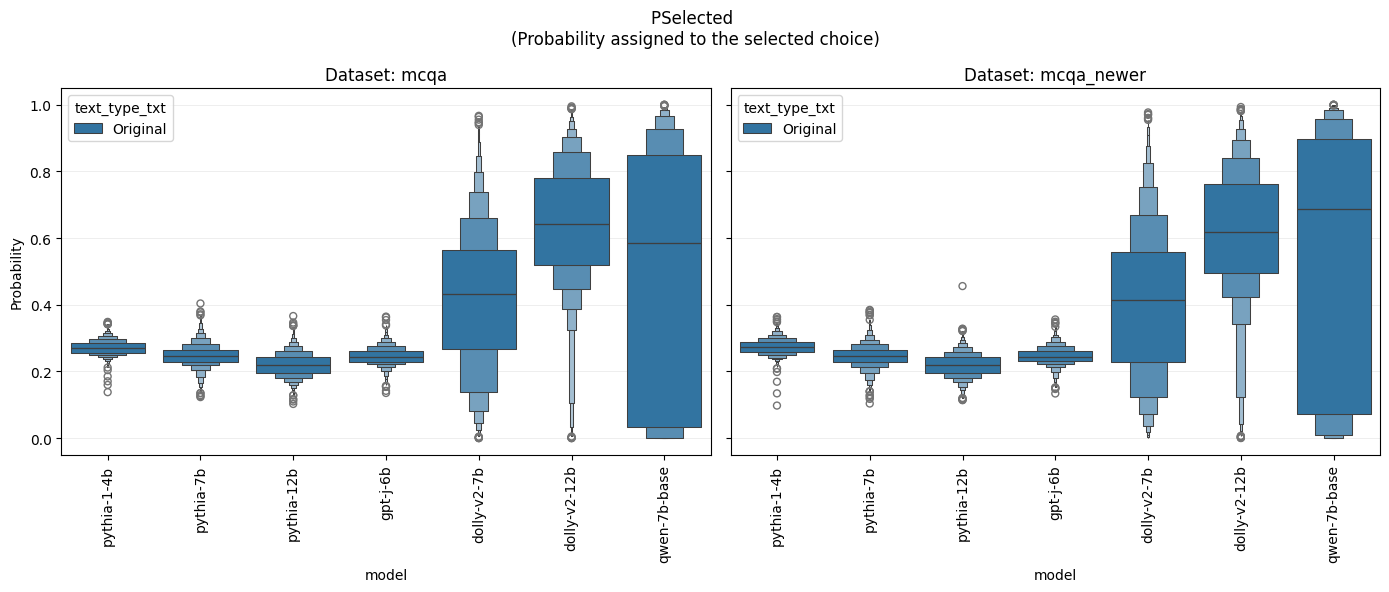}
        \caption{
            \textbf{Probabilities assigned to the selected choice.}
            Box plots showing the distribution of probabilities assigned to selected answers by different models for original questions and paraphrases. Results are shown for \textit{mcqa} (left) and \textit{mcqa\_newer} (right) datasets. Higher probabilities indicate greater model confidence in selected predictions.            
        }
        \label{fig:task2_p_selected}
\end{figure}

\begin{figure}[hbt]
    \centering
        \centering
        \includegraphics[width=1.0\linewidth]{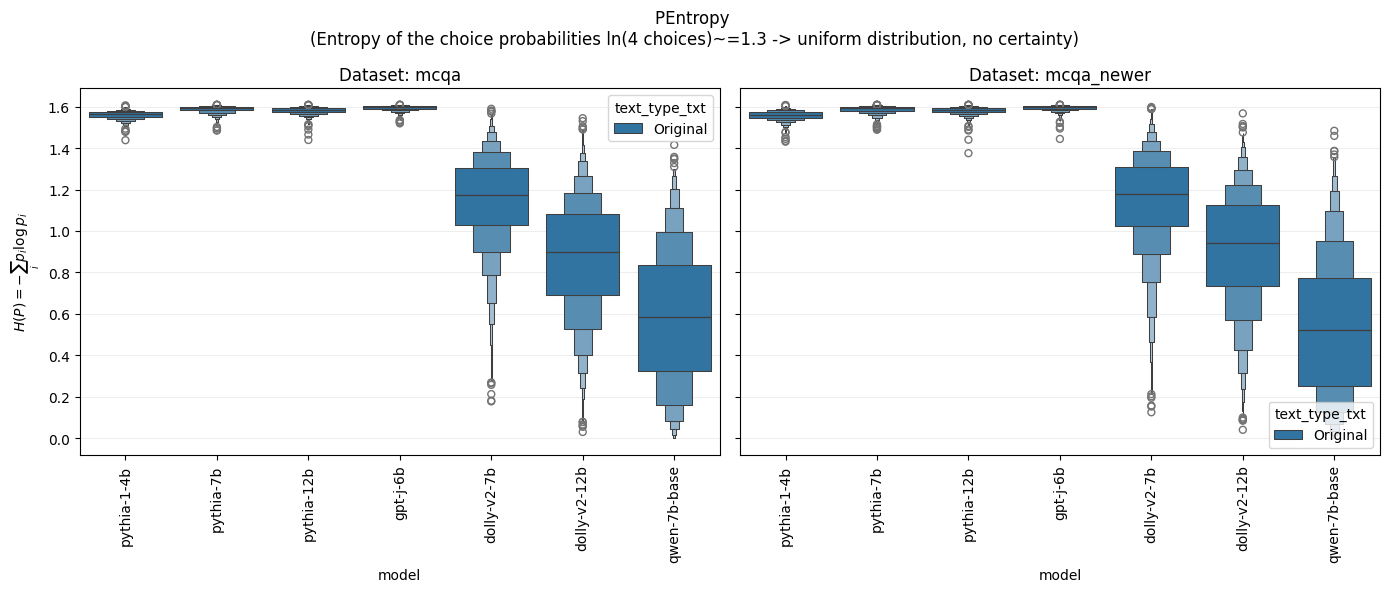}
        \caption{
            \textbf{Entropy of choice probabilities.}
            Box plots showing the distribution of entropy values across different models. Results are shown for \textit{mcqa} (left) and \textit{mcqa\_newer} (right) datasets. Higher entropy values indicate more uniform probability distributions across answer choices, reflecting greater model uncertainty. Maximum entropy of $ln(4) \approx 1.39$ corresponds to uniform distribution across four choices.
        }
        \label{fig:task2_p_entropy}
\end{figure}

We analyze results from the memorization probing task, where models choose between original sentences and semantically equivalent paraphrases. The probability (Figures \ref{fig:task2_p_true} and \ref{fig:task2_p_selected}) and entropy (Figure \ref{fig:task2_p_entropy}) analyses demonstrate no preference for original (potentially memorized) text over paraphrases, providing direct evidence against verbatim memorization as a driver of causal reasoning performance.

\end{document}